\documentclass[acmtog,natbib=true,nonacm]{acmart}
\usepackage{booktabs} 
\usepackage{amsfonts}
\usepackage{enumitem}
	
\usepackage{multirow}
\usepackage{color, colortbl}
\definecolor{gray}{gray}{0.926}
\definecolor{LightCyan}{rgb}{0.935,1,1}
\definecolor{red}{rgb}{1,0,0}
\definecolor{orange}{rgb}{1,0.9,0.8}
\definecolor{yellow}{rgb}{1,1,0.6}
\usepackage{subcaption}
\usepackage{bm}
\usepackage{algorithm}
\usepackage{algorithmic}

\usepackage{nicefrac}
\usepackage{paralist}
\usepackage{colortbl}
\newcommand\blfootnote[1]{%
  \begingroup
  \renewcommand\thefootnote{}\footnote{#1}%
  \addtocounter{footnote}{-1}%
  \endgroup
}

\newcommand{\bTheta}{\boldsymbol{\Theta}}

\newcommand{\J}{\mathbf{J}}

\newcommand{\pderiv}[2]{\frac{\partial #1}{\partial #2}}

\newcommand{\ie}{\textit{i.e. }}
\newcommand{\eg}{\textit{e.g. }}

\begin{document}

\title{Evolutive Rendering Models}

\author{Fangneng Zhan$^*$}
\affiliation{
 \institution{MPI for Informatics, VIA Research Center}
 \country{Germany}}
\email{fzhan@mpi-inf.mpg.de}
\author{Hanxue Liang$^*$}
\affiliation{
 \institution{University of Cambridge}
 \country{UK}}
\email{hl589@cam.ac.uk}
\author{Yifan Wang}
\affiliation{
 \institution{Stanford University}
 \country{USA}}
\email{yifan.wang@stanford.edu}
\author{Michael Niemeyer}
\affiliation{
 \institution{Google}
 \country{Zurich}}
\email{mniemeyer@google.com}
\author{Michael Oechsle}
\affiliation{
 \institution{Google}
 \country{Zurich}}
\email{michaeloechsle@google.com}
\author{Adam Kortylewski}
\affiliation{
 \institution{MPI for Informatics}
 \country{Germany}}
\email{akortyle@mpi-inf.mpg.de}
\author{Cengiz Oztireli}
\affiliation{
 \institution{Google}
 \country{Zurich}; \institution{University of Cambridge}
 \country{UK}}
\email{aco41@cam.ac.uk}
\author{Gordon Wetzstein}
\affiliation{
 \institution{Stanford University}
 \country{USA}}
\email{gordon.wetzstein@stanford.edu}
\author{Christian Theobalt}
\affiliation{
 \institution{MPI for Informatics, VIA Research Center}
 \country{Germany}}
\email{theobalt@mpi-inf.mpg.de}

\begin{abstract}
The landscape of computer graphics has undergone significant transformations with the recent advances of differentiable rendering models. These rendering models often rely on heuristic designs that may not fully align with the final rendering objectives. We address this gap by pioneering \textit{evolutive rendering models}, a methodology where rendering models possess the ability to evolve and adapt dynamically throughout the rendering process.
In particular, we present a comprehensive learning framework that enables the optimization of three principal rendering elements, including the gauge transformations, the ray sampling mechanisms, and the primitive organization.
Central to this framework is the development of differentiable versions of these rendering elements, allowing for effective gradient backpropagation from the final rendering objectives. 
A detailed analysis of gradient characteristics is performed to facilitate a stable and goal-oriented elements evolution.
Our extensive experiments demonstrate the large potential of evolutive rendering models for enhancing the rendering performance across various domains, including static and dynamic scene representations, generative modeling, and texture mapping.
\end{abstract}


\begin{teaserfigure}
    \includegraphics[width=1.0\textwidth]{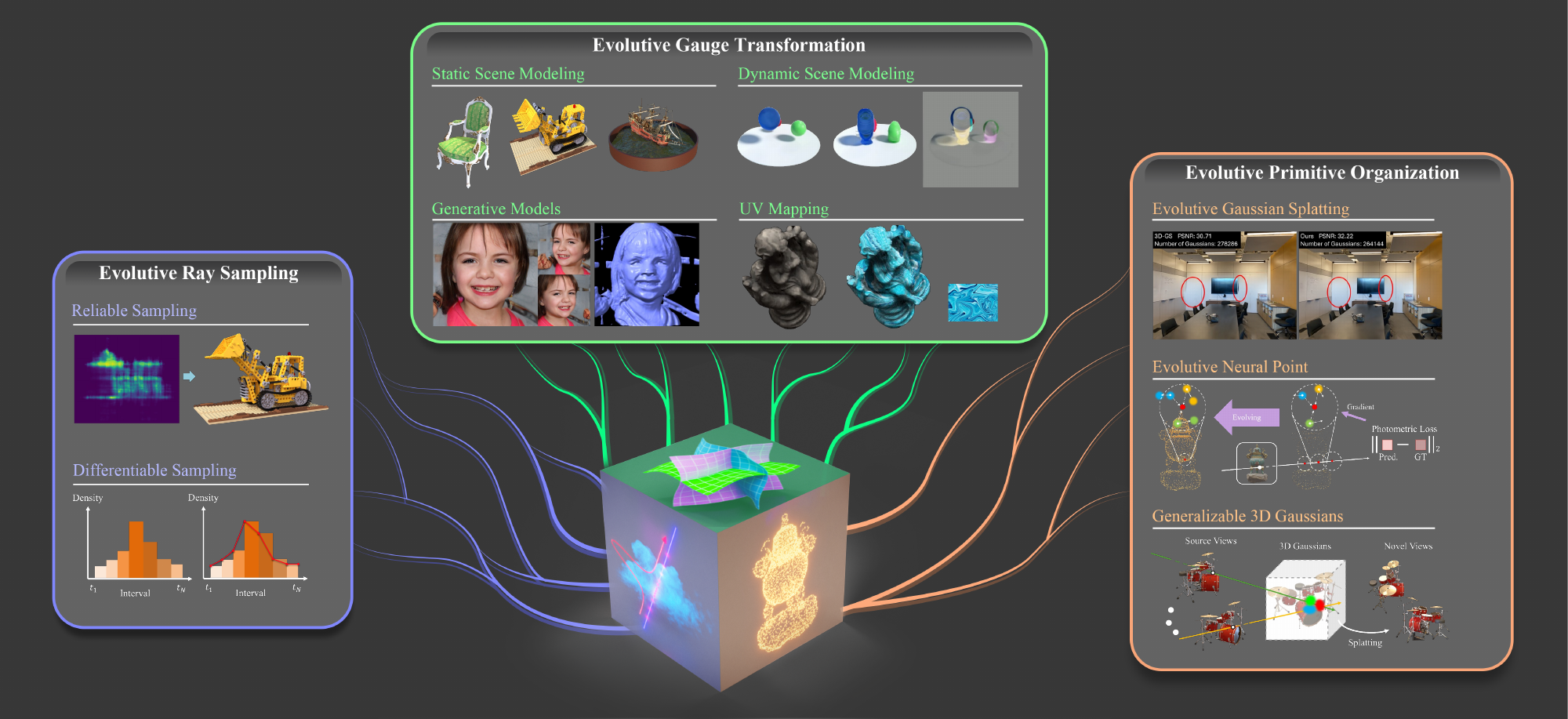}
    \caption{
    We present Evolutive Rendering Models (ERM), a framework with broad applications that enhances existing models by incorporating evolutive rendering elements, and unlocks new possibilities for previously unattainable tasks. ERM exhibits superior performance across diverse representation types (MLP-based, grid-based, and point-based models), as well as various rendering mechanism (volumetric rendering and splatting).
    }
    \label{fig_teaser}
\end{teaserfigure}

\thanks{$^*$ These authors contributed equally to this work}

\maketitle

\blfootnote{Project page: \href{https://fnzhan.com/Evolutive-Rendering-Models/}{\color{magenta}{\url{https://fnzhan.com/Evolutive-Rendering-Models/}}}}

\section{Introduction}
The field of computer graphics has undergone a remarkable revolution with the advancement of differentiable rendering models highlighted by representative works~\cite{mildenhall2020nerf,muller2022instant,kerbl20233d}.
Such models are essential for accurate environment digitization and immersive XR experiences, and are increasingly relevant across a diverse array of industries including construction, entertainment, and robotics. It bridges the gap between real-world data acquisition and digital visual representation. The versatility and applicability of differentiable rendering in these domains underscore its significance as a transformative tool in modern graphics and machine learning applications.

Despite these advancements, a fundamental challenge persists in the development of rendering models: the reliance on one-fits-all, heuristic, hand-defined rules. These heuristics, while beneficial in offering a starting point for model design, often compromise the expressiveness and adaptability of the models. They tend to impose rigid constraints, limiting the capacity of these models to adapt and evolve in response to diverse and dynamically changing optimization state and rendering objectives.

In this work, we introduce \textbf{E}volutive \textbf{R}endering \textbf{M}odels. The ERM is designed to evolve autonomously towards more optimal states, offering an alternative to traditional heuristic and rule-based methods. 
This work focuses on three prevalent elements in scene representation and rendering: (1) a gauge transformation \cite{zhan2023general}, denoting the conversion between distinct measuring systems, to perform space mapping to index radiance fields, (2) a sampling mechanism to perform ray sampling for volume rendering, and (3) a primitive organization in the space to perform point-based rendering. 
Central to this approach is the employment of differentiable version of above elements, which paves the way for a fully learnable system, adaptively guided by gradient-based optimization. 
Anchoring this innovative framework is a principled optimization paradigm termed as \textbf{relay learning mechanism}, meticulously devised through rigorous gradient analysis. This relay learning mechanism ensures robust and stable evolution of the rendering models across diverse test scenarios, marking a significant stride in the field.

Aligned with the three rendering elements, we include several concrete samples to demonstrate its potential applications:
\begin{inparaenum}
\item Evolutive Gauge Transformation: we develop a parametric mapping technique between Euclidean 3D space and low-dimensional 2D space \cite{chan2022efficient,fridovich2023k,zhan2023general}. This technique employs learnable components for dynamic adaptation of the mapping process, facilitating flexible and expressive scene representation.
\item Evolutive Ray Sampling: we propose a gradient-guided sampling strategy to improve the efficiency of ray-marching techniques. This strategy notably enhances reconstruction quality in NeRF-based volumetric rendering~\cite{muller2022instant,sun2022direct}, promoting the rendering efficiency and overall quality in current methodologies.
\item Evolutive Primitive Organization: in the realm of point-based rendering, our approach innovates by incorporating a learnable component to optimize the densification and pruning phases~\cite{xu2022point,kerbl20233d}. This adaptation is responsive to the current optimization state, thus improving both precision and efficiency.
\end{inparaenum}
The practical applications of this approach range from promoting performance of existing models by incorporating evolutive rendering elements, to unlocking new possibilities for tasks that were unattainable with existing models.
Our experiments are performed over a variety of representation types (MLP-based, grid-based, and point-based models), as well as different rendering types (volumetric rendering and splatting), highlighting the adaptability and broad applicability of our evolutive rendering approach.

In summary, our key contributions are:
\begin{compactitem}
\item Introduction of the Evolutive Rendering Model (ERM) for autonomous evolution towards optimal rendering states;
\item Integration of differentiable components as an alternative to heuristic and rule-based elements, facilitating a fully learnable system;
\item Establishing a relay learning mechanism, rigorously grounded in gradient analysis, to facilitate robust and stable evolution in a myriad of rendering applications.
\item Demonstration of the superiority of ERM through concrete examples within contemporary computer graphics, which span over a variety of representation types and rendering techniques.
\end{compactitem}
\begin{figure*}[t]
    \includegraphics[width=1.0\linewidth]{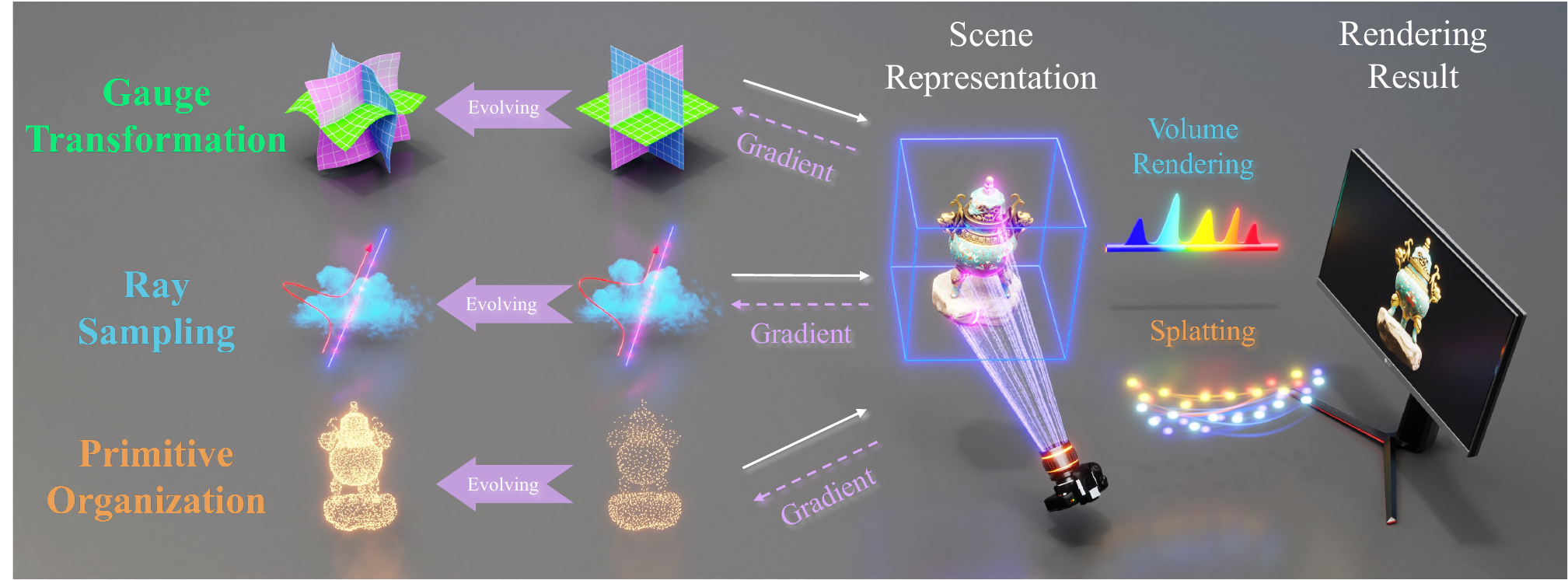}
    \caption{Evolutive Rendering Models covers three principal rendering elements: gauge transformation, ray sampling and primitive organization. All three elements can be applied in volume rendering, while splatting only employs evolutive gauge transformation and primitive organization.}
    \label{fig_framework}
\end{figure*}

\hspace{5pt}

\section{Related Work}

Recent work has significantly advanced the field of scene representation, particularly through the development of NeRF \& 3DGS \cite{mildenhall2020nerf,kerbl20233d} and their various extensions. These advancements have found widespread application in numerous areas of vision and graphics, notably in view synthesis \cite{yu2021plenoctrees,lindell2021autoint,fridovich2022plenoxels,reiser2021kilonerf,sun2022direct}, generative models \cite{schwarz2020graf,niemeyer2021giraffe,chan2021pi}, and surface reconstruction \citep{wang2021neus,oechsle2021unisurf,yariv2021volume}.

In contrast to the aforementioned application works, our proposed evolutive rendering models cater to the fundamental elements in scene representation and rendering, including gauge transformation, ray sampling, and primitive organization.

\subsection{Gauge Transformation}

Under the context of neural rendering, gauge transformation denotes the mapping between two coordinate systems. 
This concept is correlated with the prevailing paradigm of learning deformation for dynamic modeling \citep{tretschk2021non,tewari2022disentangled3d,park2021nerfies, pumarola2020dnerf,peng2021animatable}, which actually learns a mapping within one coordinate system.
A diverse array of gauge transformations has been extensively investigated in neural fields, serving various objectives like efficient rendering \cite{chan2022efficient,chen2022tensorf,muller2022instant,zhan2023general}.
A pre-defined mapping function is usually employed as the gauge transformation. 
A typical example is orthogonal mapping, which involves projecting a 3D space onto 2D planes as in \cite{chan2022efficient,peng2020convolutional}.
Expanding this concept, TensoRF \cite{chen2022tensorf} propels this research forward for fast and efficient optimization by projecting the 3D scene space into 2D planes and 1D vectors; \citet{cao2023hexplane,fridovich2023k} propose to project 4D space onto 2D planes for dynamic modeling.

Concurrently, some studies also ventured into learning the gauge transformation for specialized tasks within neural fields. NeuTex \cite{xiang2021neutex} and NeP \cite{ma2022neural}, for instance, focus on learning the mapping from 3D points to 2D texture spaces.
Neural Gauge Fields \cite{zhan2023general} generally explores the problem of gauge transformation and its optimization.
However, all above works necessitate certain regularizations to facilitate stable optimization, which is a cumbersome process and hinders it for practical applications.
In this work, we introduce a relay learning mechanism which allows efficient optimization of gauge transformations without any regularizations, unlocking its potential for various gra-phic applications.

\subsection{Ray Sampling}

Ray sampling is pivotal in promoting the efficiency of volume rendering. The original NeRF \cite{mildenhall2020nerf} employs a coarse-to-fine sampling strategy, selecting points based on their contribution to the final rendering.

However, the coarse sampling phase entails a cumbersome process of querying radiance fields per point along a ray. To address this, a series of work focuses on directly generating target samples for a given ray. NeuSample \cite{fang2021neusample} suggests that the coarse stage can be substituted with a lightweight module parameterized by an MLP. Similarly, DONeRF \cite{neff2021donerf} and TermiNeRF \cite{piala2021terminerf} propose replacing vanilla NeRF's coarse sampling with a sampling network that predicts object surface depths. Yet, these methods hinge on the availability of depth maps, constraining their practical utility. In scenarios without depth priors, AdaNeRF \cite{kurz2022adanerf} introduces a sampler network that converts rays into discrete probabilities, albeit involving a complex optimization procedure. ProNeRF \cite{bello2023pronerf} opts for estimating sampled points in a coarse-to-fine manner, supplemented by multiview projection to capture geometric information. Overall, above approach, by sidestepping the structure of radiance fields, is prone to issues like geometry collapse and overfitting.

Conversely, another research trajectory maintains the coarse-to-fine sampling paradigm. NeRF in Detail \cite{arandjelovic2021nerf} conducts initial coarse sampling of a ray, followed by a network that refines target points from the coarsely sampled points' features. MipNeRF 360 \cite{barron2022mip} suggests distilling information from the density field into a sampling field for ray sampling. However, the distillation loss employed is based on heuristics, presupposing an alignment between the sampling and density fields. Rather than relying on heuristic designs, RVS \cite{morozov2023differentiable} enables the gradient from the training objective to optimize the sampling field. Nevertheless, this method is primarily applicable to MLP-based radiance fields. In our work, we demonstrate the utility of training objective gradients for general representation types of sampling fields, employing a straightforward yet efficient strategy known as relay learning mechanism for optimization.

\subsection{Point-based rendering}
While neural radiance fields (NeRF)~\cite{mildenhall2020nerf} are prevalent, as highlighted in prior work, point-based primitives offer an efficient alternative, leveraging GPU hardware rasterization~\cite{pineda1988parallel}. 
Traditional single-pixel point rendering faces challenges like holes in sparse point clouds~\cite{catmull1974subdivision, schaufler1998per}, addressed in part by convolutional neural networks~\cite{aliev2020neural}. However, these methods struggle with view consistency and generalization.
The rise of NeRF has led to techniques like Point-NeRF~\cite{xu2022point}, combining points with volumetric rendering, albeit at a loss of rasterization efficiency. 
Notably, EWA splatting~\cite{zwicker2002ewa} interprets point rendering through a signal reconstruction lens, employing Gaussian reconstruction kernels to reconstruct continuous signals from discrete samples.
Recent advancements have focused on integrating differentiability into point-based rasterizers~\cite{yifan2019differentiable, wiles2020synsin, lassner2021pulsar, muller2022unbiased, kerbl20233d}. Notably, Kerbl et al. have developed an efficient differentiable point rasterizer, synergizing EWA Splatting with volume rendering. This innovation facilitates both rapid scene reconstruction and photorealistic novel-view synthesis in real-time.

Optimizing point-based inverse rendering models crucially depends on the organization of primitives. A common strategy involves periodic resampling through splitting and pruning, essential for optimization stability~\cite{zheng2023pointavatar,kerbl20233d}. However, current splitting and pruning strategies are heuristic and not optimized concurrently. PixelsSplat~\cite{charatan2023pixelsplat} critiques these strategies, proposing a differentiable parameterization of Gaussian primitives less prone to local minima. In our work, we introduce an innovative primitive organization procedure, integrating differentiable splitting and pruning within an evolutionary optimization framework.

\hspace{5pt}
\section{Methodology}
\label{methodology}

\begin{figure}[ht]
    \includegraphics[width=1.0\linewidth]{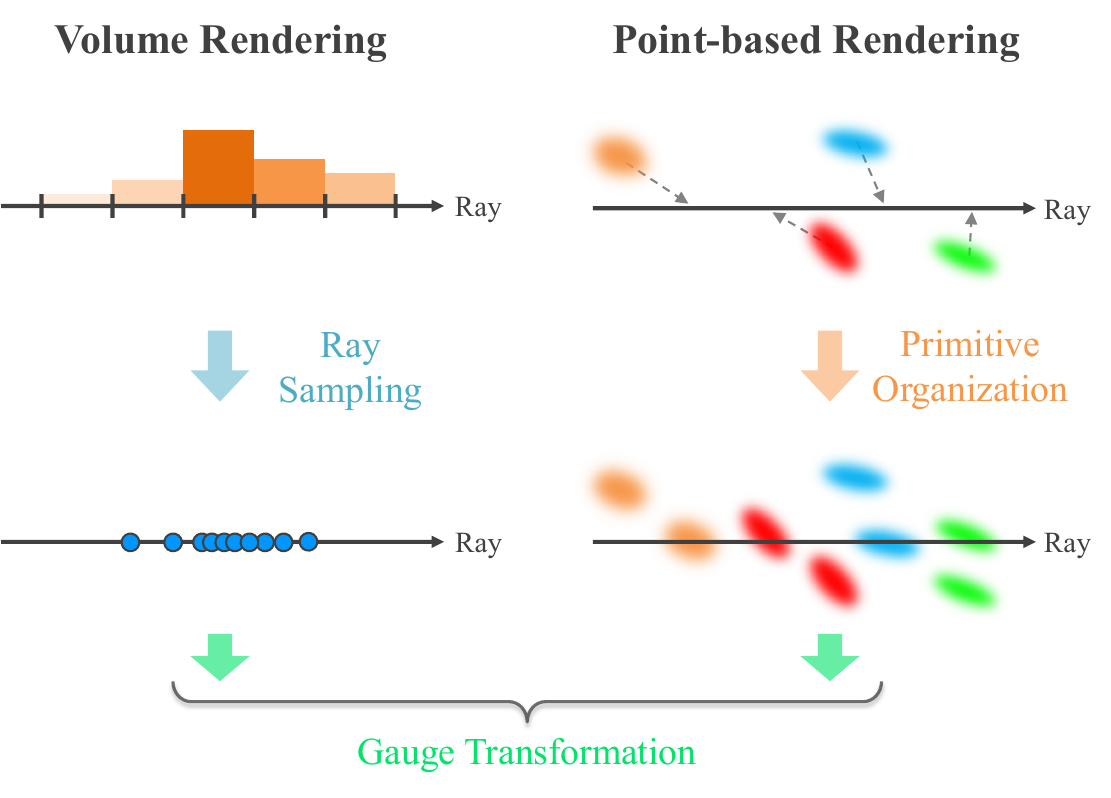}
    \caption{
        A conceptual illustration of rendering elements, including ray sampling in volume rendering, primitive organization in point-based rendering, and gauge transformations.
        Volumetric and point-based rendering can be performed in a unified manner: accumulating or blending discrete points relevant to the given ray.
        }
    \label{fig_concept}
\end{figure}

This section describes the methodology underlying the proposed Evolutive Rendering Model (ERM).
As illustrated in Fig.~\ref{fig_framework}, our framework covers three principal rendering elements including the gauge transformation, the ray sampling, and the primitive organization.
The gauge transformation can be performed to transform discrete points to another coordinate system to index scene representation as shown in Fig. \ref{fig_concept}.
Notably, ray sampling (in volume rendering) and primitive organization (in splatting) share essentially the same key operation in rendering pipelines, i.e., yielding desired discrete positions in the continuous space to perform rendering.
The unified formulation of volume rendering and point-based (or splat-based) rendering can be written as:
\begin{equation}
    C = \sum_{i \in \mathcal{N}} c_i \alpha_i \prod_{j=1}^{i-1} (1-\alpha_j),
\label{eq_rendering}
\end{equation}
where $C$ is the color of a image pixel, $c_{i}$ is the color of discrete point in the space.
In volume rendering, $\alpha_i$ can be computed according to the point density $\sigma$ as $\alpha_i = (1-exp(-\sigma_i \delta_i))$;
in point-based rendering, $\alpha_i$ is given by evaluating a 2D Gaussian according to its covariance and per-point opacity.

As two different lines of research, ray sampling and primitive organization are incompatible in rendering models for most cases \footnote{PointNeRF \cite{xu2022point} is a special case which can employs both ray sampling and primitive organization.}.
For instance, volume rendering models can only employ gauge transformation and ray sampling, while the point-based rendering models can only employ gauge transformation and primitive organization.

\subsection{Evolutive Gauge Transformation}

Generally, a gauge defines a measuring system, e.g., pressure gauge and temperature gauge.
In the context of neural rendering, a measuring system (i.e., gauge) is a specification of parameters to index a radiance field,
e.g., 3D Cartesian coordinate in original NeRF~\cite{mildenhall2020nerf}, triplane in EG3D~\citep{chan2022efficient}, plane \& vector in TensoRF \cite{chen2022tensorf}.
The transformation between different measuring systems is referred as \textbf{Gauge Transformation}.
In radiance fields, gauge transformations are defined as the transformation from the original space to another gauge system to index radiance fields. This additional transform could introduce certain bonus to the rendering, e.g., low memory cost, high rendering quality, or explicit texture, depending on the purpose of the model.

Typically, the gauge transformation is performed via a pre-defined function, e.g., an orthogonal mapping in 3D.
This pre-defined function is a general design for various scenes, which means it is not necessarily the best choice for a specific target scene.
Moreover, it is a non-trivial task to manually design an optimal gauge transformation which aligns best with the complex training objective.
We thus introduce the concept of \textbf{E}volutive \textbf{G}auge \textbf{T}ransformation (\textbf{EGT}) to optimize a desired transformation directly guided by the final training objective.

The gauge transformation can be parameterized by an MLP-network \& feature grid, or per-point property for the case of point-based rendering.
For a point $x \in \mathbb{R}^3$ in the original space, the evolutive gauge transformation outputs the corresponding coordinate in the target space. 
The output coordinate can be the absolute value or a residual offset.
On the other hand, the efficient optimization of EGT is a challenging task, previous works~\cite{zhan2023general} regularize the optimization process for implicit fields, which however is too heavy for practical applications.
We thus introduce optimization strategies without clearly slowing down the training speed.

\begin{figure}[ht]
    \includegraphics[width=1.0\linewidth]{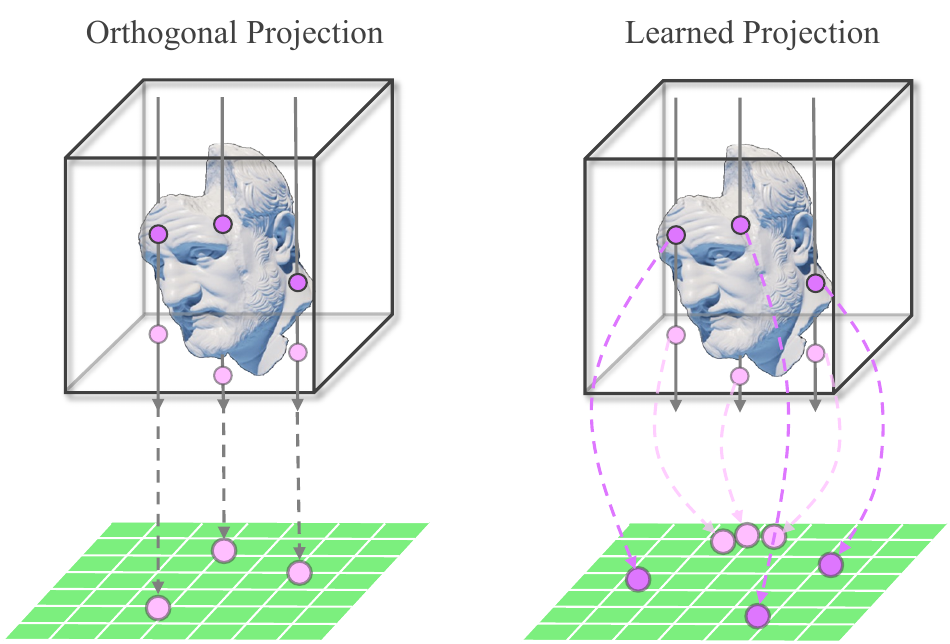}
    \caption{
    Motivation of evolutive gauge transformation. In this example, instead of mapping the 3D euclidean space to the 2D plane by orthogonal projection (left), we learn a more flexible and adaptive mapping (right).
        }
    \label{fig_gauge}
\end{figure}

\begin{figure}[ht]
    \includegraphics[width=1.0\linewidth]{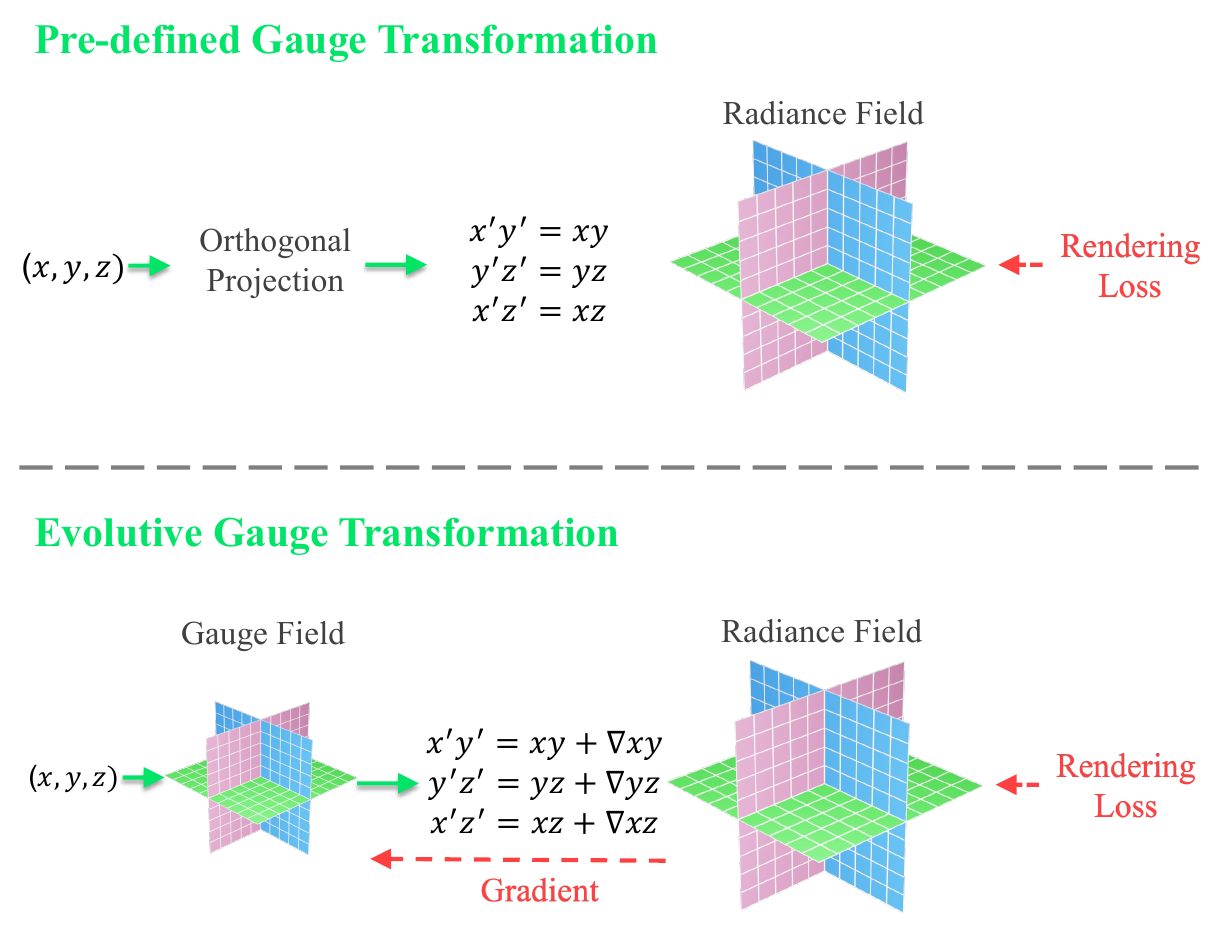}
    \caption{
        The illustration of predefined (upper) and evolutive (lower) gauge transformation. We implement the evolutive transformation by predicting an offset to the pre-defined transformation.
        }
    \label{fig_gauge_compare}
\end{figure}

Generally, the gradient is unstable at the initial training stage, which is especially the case for grid-based representation as analyzed in Sec. \ref{sec_analysis}.
We thus introduce a deferred learning strategy for stable optimization of gauge transformation.
As shown in Fig.~\ref{fig_gauge_compare}, we train the model with a pre-defined gauge transformation (\eg, orthogonal projection) at the initial stage. The gradient will become more stable when a coarse scene representation is learned. Thus, for the later stage, we replace the predefined gauge transformation with the learnable counterpart, and jointly optimize it with the scene representation.
The training strategy can help to stabilize the training and accelerate the convergence for general representations, especially for grid-based representation.

\subsection{Evolutive Ray Sampling}

In volume rendering process, densely evaluating the radiance field network at query points along each camera ray is inefficient, as only few regions contribute to the rendered image.
Thus, a coarse-to-fine sampling strategy~\cite{mildenhall2020nerf} is usually employed to increase rendering efficiency by allocating samples proportionally to their expected effect on the final rendering.
To achieve coarse-to-fine sampling, a sampling field is included in the rendering pipeline.
At first, a set of points are uniformly sampled along a ray $[t_1, \cdots, t_N]$, to evaluate the sampling field.
For piecewise constant approximation, point density within each bin $t_{i} \leq \hat{t}_i \leq t_{i + 1}$ are approximated with constant density of $\sigma_i)$.
The evaluation of $[t_1, \cdots, t_N]$ yields a discrete distribution of density along the ray, which gives the color weights $w_i$ of different point as:
\begin{equation}
\label{eq_weight}
    w_i = \alpha_i \prod_{j=1}^{i-1} (1-\alpha_j) \quad  \alpha_i = (1-exp(-\sigma_i \delta_i))
\end{equation}
The color weights are normalized as $w_i = \nicefrac{w_i}{\sum_{j=1}^{N_c} w_j}$ to produce a piecewise-constant probability density function (PDF) along the ray.
According to the PDF and corresponding CDF, a second set of locations are sampled from this distribution using inverse transform sampling, which allocates more samples to more visible regions.

\begin{figure}[ht]
    \includegraphics[width=1.0\linewidth]{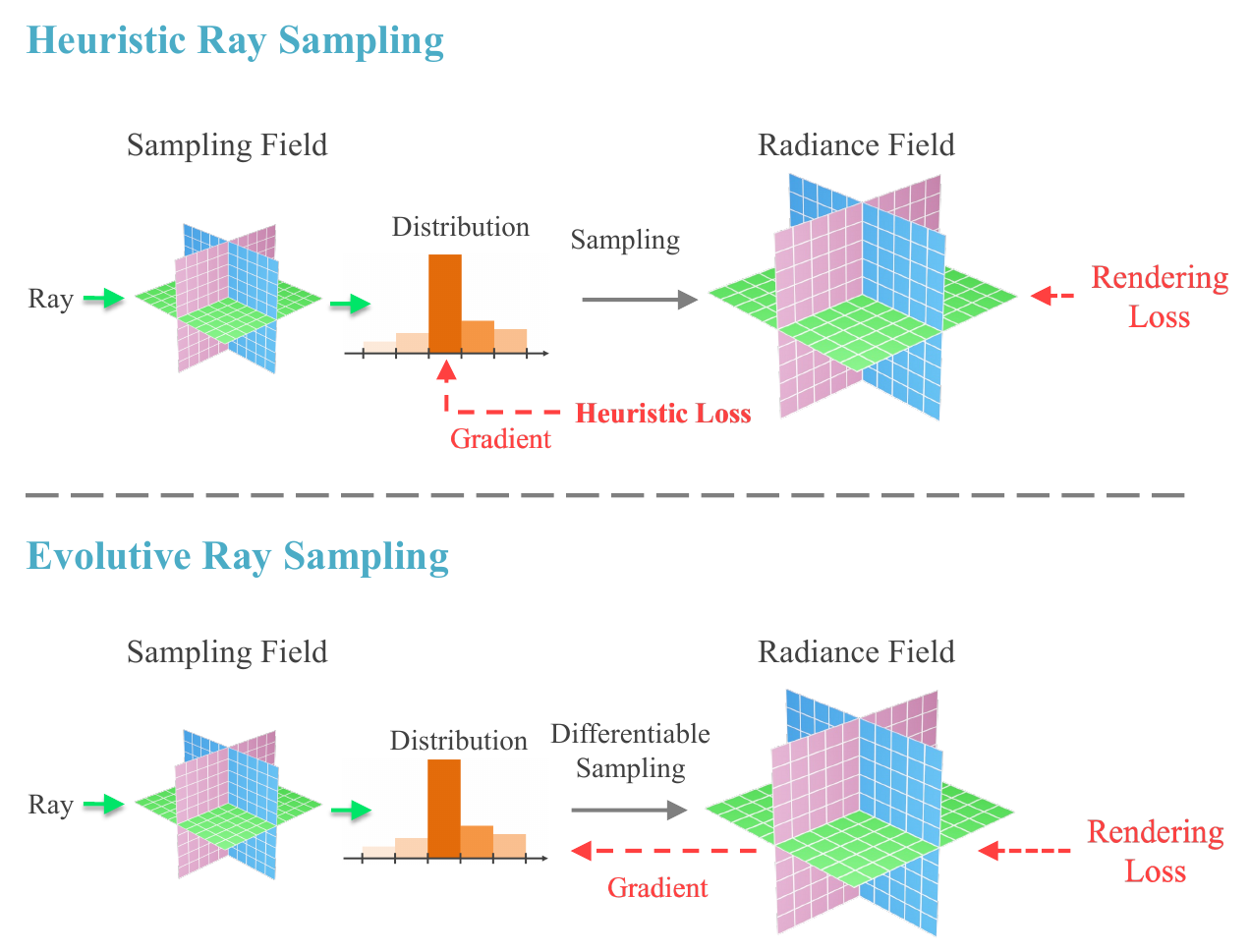}
    \caption{
        The illustration of pre-defined ray sampling and evolutive ray sampling.
        }
    \label{fig_sampling_compare}
\end{figure}

To optimize the sampling fields, previous work either treat it as a radiance field trained with photometric loss or distilling the density knowledge from the radiance fields as shown in Fig. \ref{fig_sampling_compare}.
Thus, all of them are making a heuristic assumption: the best of sampling fields should be aligned with the density fields.
However, the objective of sampling field is to select the best set of points for the evaluation of radiance field, while the density field aims to yield the best rendering results.
Thus, previous heuristic assumption will bias the optimization objective of sampling fields.
On the other hand, it is non-trivial to manually design the training objective for the sampling fields, which should be determined by the radiance fields as sampling fields serve for radiance fields.
To this end, we propose to backpropagate gradients from the radiance fields (\ie, rendering loss) to optimize the sampling field directly, eliminating the need to heuristically design auxiliary loss supervision.
However, for this case of piecewise constant approximation, the CDF is a discontinuous step function, which hinders the gradient backpropagation in the sampling process.

\begin{figure}[h]
    \includegraphics[width=1.0\linewidth]{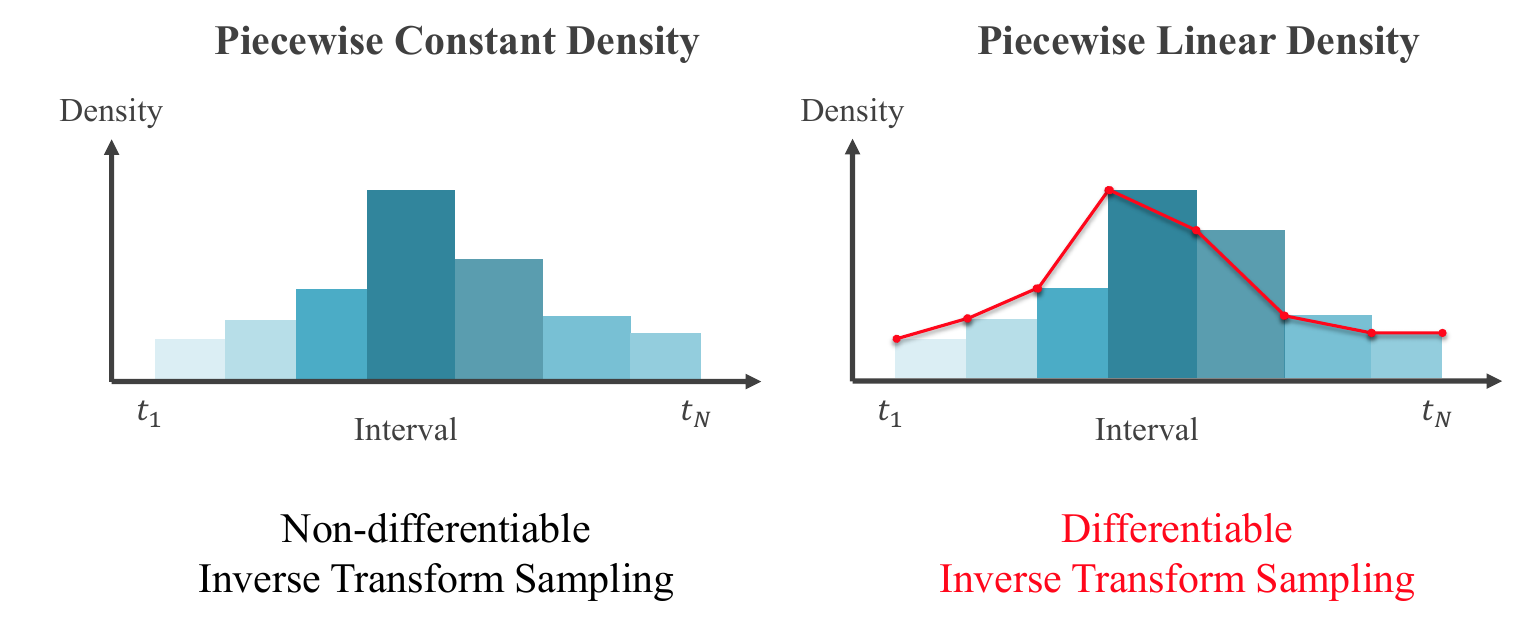}
    \caption{
    Differentiable sampling with piecewise linear approximation.
        }
    \label{fig_sampling_linear}
\end{figure}

To achieve differentiable sampling, we adopt a piecewise linear density to approximate the opacity \cite{morozov2023differentiable,uy2023nerf}, as illustrated in Fig.~\ref{fig_sampling_linear}.
Specifically, we compute $\sigma(t), t \in [t_i, t_{i+1}]$ by interpolating the values between the interval points $t_i$ and $t_{i+1}$:
\begin{equation}
    \sigma (t'_i) = \sigma_{i+1} \frac{t_{i+1} - t}{t_{i+1} - t_i} + \sigma_i \frac{t - t_i}{t_{i+1} - t_i}.
\end{equation}
Given these piecewise linear approximations of $\sigma_i, i \in [1, N]$, we can yield a continuous PDF and CDF according to Eq. (\ref{eq_weight}).
With the continuous CDF, the sampling process with inverse transform is differentiable function with respect to the density field $\sigma_i$ and can back-propagate the gradients to optimize the sampling fields.

This differentiable sampling algorithm can be smoothly integrated into the hierarchical sampling scheme originally proposed in NeRF. Here we do not change the final color approximation, utilizing the original one, but modify the way the coarse density network is trained. The method we introduce consists of two changes to the original scheme. Firstly, we replace sampling from piecewise-constant PDF along the ray defined by weights $w_i$ with differentiable sampling algorithm that uses piecewise linear approximation of $\sigma_{\bm{r}}$ and generates samples from $p_r(t)$ using inverse CDF. Secondly, we remove the auxiliary reconstruction loss imposed on the coarse network. Instead, we propagate gradients through sampling. This way, we eliminate the need for auxiliary coarse network losses and train the network to solve the actual task of our interest: picking the best points for evaluation of the fine network. All components of the model are trained together end-to-end from scratch.

\subsection{Evolutive Primitive Organization}\label{sec:EPO}
\begin{figure}[ht]
\centering
\begin{subfigure}[b]{0.5\textwidth}
   \includegraphics[width=0.95\linewidth]{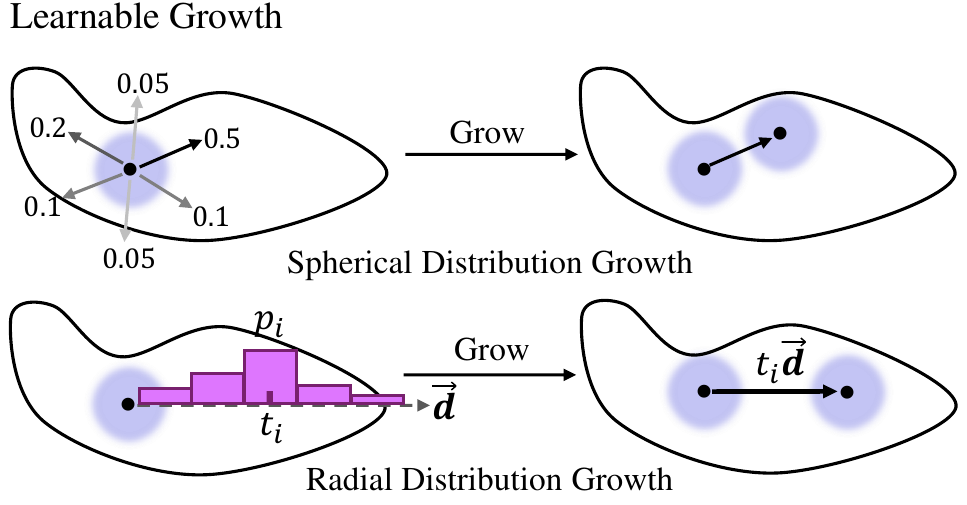}
   \caption{}
   \label{fig:poe1} 
\end{subfigure}

\begin{subfigure}[b]{0.5\textwidth}
   \includegraphics[width=0.95\linewidth]{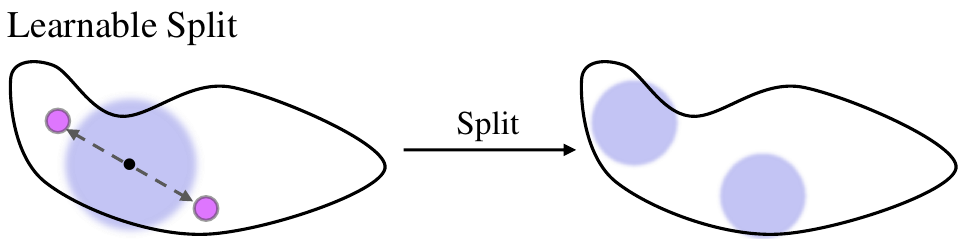}
   \caption{}
   \label{fig:poe2}
   
\end{subfigure}

\caption[Two numerical solutions]{An illustration of evolutive primitive organization. In fig. (a),  we elucidate evolutive primitive growth where we consider two fundamental forms of primitive growth distribution: spherical distribution growth has a pre-difined growth length and learn growth direction probability; radial distribution growth assums known growth direction and learn growth length probability. In fig. (b), we show evolutive primitive split, where a split shift term is learned to decide the location of newly-split primitives.}
\label{fig:poe}
\end{figure}

Point-based representations employ a set of geometric primitives (e.g. neural points in Point-NeRF \cite{xu2022point}, Gaussians splats in 3D-GS \cite{kerbl20233d}) for scene rendering. These primitives compose of attributes that encode the geometry and radiance field, which can be rendered and optimized via volume render or rasterization operation. They are usually initialized from SFM and optimizing their attributes directly via gradient descent suffers from local minima. Previous techniques try to alleviate this issue in per-scene fitting by employing pre-defined optimization heuristics, such as point growing and pruning in Point-NeRF and Adaptive Density Control in 3D-GS. However, these heuristic operations can be sub-optimal because they are non-differentiable and may misalign with final training objective. Furthermore, their non-differentiable nature also impedes their applicability in cross-scene generalizable settings. We thus propose primitive organization evolution, where scene primitives will be implicitly grown and split during training while maintaining gradient flow directly from training objective.  Our proposed approach not only overcomes the challenges posed by non-differentiability but also facilitates the extension of current techniques to feed-forward generalizable settings, which we will elaborate in Section \ref{generalizable-gs}. An illustration of evolutive primitive organization is shown in Fig.~\ref{fig:poe}.

We denote the position of existing scene primitives as $\mu_k \in \mathbb{R}^3$. When primitive organization evolve to grow new primitives (eg. new primitives growth in under-reconstructed region),
we learn a grown term $\delta \mu_k=td \in \mathbb{R}^3$ for the emergent primitives, where $d\in \mathbb{R}^3$ is the direction of grown term and $t\in \mathbb{R}$ is its length. The location $\mu'_k$ of new primitives will be $\mu_k+\delta \mu_k$. The newly-grown primitives will be rendered during each iteration, which allows the grown term to be optimized by the final training objective throughout whole optimization process.
Unfortunately, we find directly regressing grown term makes the training unstable, which is susceptible to local minima. Instead, we consider two most fundamental forms of primitive growing distribution: spherical distribution and radial distribution, elucidated in Fig.~\ref{fig:poe1}. In the context of spherical distribution growth, nascent primitives expand along a sphere enveloping the original primitives, with the growth length $t$ predefined and growth direction $d$ being learned. Conversely, in radial distribution growth, the grow direction is predefined, while the extent of growth $t$ along this direction will be learned. The combination of these two primary grow distribution actually spans the entirety of the potential growth space.
 
To stabilize the learning process, we choose to learn these two forms of distribution in discrete space. In more details, when to learn the spherical distribution of primitive growth, we first pre-define a set of $N$ uniformly distributed potential growing directions $\left\{d_1,d_2, ..., d_N\right\}$, and each emergent primitive will learn a probability distribution of grow directions $Q \in \mathbb{R}^N$, where its i-th element $q_i$ represent the probability of growing along direction $d_i$. The actually grow direction $d$ is chosen to be the direction with maximum probability:
\begin{equation}
    d=d_i, i=argmax(Q)
\end{equation}
As Argmax operations
is non-differentiable, we apply reparameterization trick by
replacing Argmax operation with Softmax in gradient back-propagation. The pseudo
code of the forward \& backward propagation of the grow primitives is given in Algorithm \ref{alg1}.
Similarly, for its counterpart of radial distribution growth, given grow direction $d$, we learn the growth distance by predicting the probability that new primitive will exist at distance $t$ along the direction. We discretize the extention along the direction into $N$ bins with distance $\left\{t_1,t_2, ..., t_N\right\}$. And we learn a discrete grow distance probability $Q \in \mathbb{R}^N$, where $q_i$ represent the probability of growing with distance $t_i$. The pseudo code is also included in Algorithm \ref{alg1}. Please note that although these two primary grow distribution forms can span the entire potential growth space, experimentally in most cases we only need to employ one of them depending on the applications.

\begin{algorithm}[htb] 
    \caption{Pseudo-code of forward \& backward propagation in primitive growth spherical/radial distribution optimizatation} 
    \label{alg1} 
    \begin{algorithmic} 
        \STATE \textbf{Input:}\\
        potential grow directions $D=\left\{d_1,d_2, ..., d_N\right\}$/potential grow distance $T=\left\{t_1,t_2, ..., t_N\right\}$\\ grow direction/distance probability $Q =[p_1, p_2, ..., p_N]$
        \\
        \textbf{Forward propagation:}\\
            \quad 1. index = Argmax($Q$)\\
            \quad 2. index-hard = One-Hot(index)\\
            \quad 3. grow direction $d$ = Matmul(index-hard, $D$)/grow distance $t$ = Matmul(index-hard, $T$)\\
        \textbf{Backward propagation:}\\
            \quad 1. index-soft = Softmax($Q$)\\
            \quad 2. grow direction $d$ = Matmul(index-soft, $D$)/grow distance $t$ = Matmul(index-soft, $T$)\\
    \end{algorithmic}
\end{algorithm}

In addition to learned primitive growth, there are circumstances that require splitting the primitives. For example, 3D-GS propose a splitting operation that splits large primitives in over-reconstructed region into two smaller ones by dividing their scale with a pre-defined scaling factor of 1.6, and initialize their location by using the original 3D Gaussian as a PDF for sampling. This whole sampling process is no-differentiable and the splitting operation won't directly align with optimization objective. Thus we additionally learn a differentiable splitting strategy by predicting the new position of split primitive (illustrated in Fig.~\ref{fig:poe2}). Similarly to learned growing operation, we learn a split mean shift $\delta\mu$ and the position of two newly split primitive will be $\mu+\delta\mu$ and $\mu-\delta\mu$ separably. The new primitive will participate in rendering process within each training iteration, allowing gradient flow to update the shift term.

\section{Optimization}
\label{sec_analysis}

For clarity, we denote the gauge transformation, the ray sampling, and the primitive organization as $\mathcal{T}$, $\mathcal{S}$, and $\mathcal{O}$, respectively.
The optimization of above rendering elements relies on the gradients derived from rendering models.
There are two main paradigms for rendering, including (1) sampling discrete points in the space to perform volume rendering (via point accumulation),
(2) organizing discrete primitives in the space to perform splat-based rendering (via $\alpha$-blending). 
Given $\mathbf{u_i}=[c_i, \sigma_i]$ and the unified formulation of volume rendering and point-based rendering
$C = \sum_{i \in \mathcal{N}} c_i \alpha_i \prod_{j=1}^{i-1} (1-\alpha_j)$, the function of per-point color contribution in can be written as $G(\mathbf{u_i})=c_i \alpha_i \prod_{j=1}^{i-1} (1-\alpha_j)$.
Note that $G$ is a differentiable function without learnable parameters.
In the next subsections, we will discuss the gradient characteristics for element optimization in volume rendering and point-based rendering, respectively.

\subsection{Gradient in Volume Rendering}

Typical volume rendering is associated with a continuous representation of the scene, necessitating point sampling mechanism $\mathcal{S}$ to yield discrete samples (with optional gauge transformation $\mathcal{T}$).
For clarity, we denote the joint process of ray sampling and gauge transformation as $\mathcal{ST}(r; \bTheta_{st})$, where $\bTheta_{st}$ and $r$ are the ray sampling \& gauge transformation parameters and a given ray.
Then, the process to yield a discrete point $p_i$ along ray $r$ can be written as $p=\mathcal{ST}(r; \bTheta_{st})$.
The discrete point $p$ is further used to query the scene representation to yield color \& density $\mathbf{u} = [c, \sigma]=f(p; \bTheta_f)$, where $f$ and $\bTheta_f$ are the representation function and parameters, e.g., MLP in implicit neural fields or feature grid in explicit neural fields.
Thus, the color contribution from $p$ can be formulated as:
\begin{equation}
\label{eq_vr}
    G(\mathbf{u})=G(f(p; \bTheta_f)) = G(f(\mathcal{ST}(r; \bTheta_{st}); \bTheta_f ) ).
\end{equation}
The gradient of $\J$ of the color contribution with respect to $\bTheta_{st}$ can be derived as:
\begin{align}
    \J = \pderiv{G(\mathbf{u})}{\mathbf{u}} 
         \pderiv{f(p; \bTheta_f)}{p} 
         \pderiv{\mathcal{ST}(r; \bTheta_{st}) }{\bTheta_{st}}.
\end{align}
As the gradient term $\pderiv{G(\mathbf{u})}{\mathbf{u}} $ is obviously stable, the optimization depends on the terms $f$ and $\mathcal{ST}$ which will be analyzed according to their parameterization types in ensuing paragraphs.

\subsubsection{Explicit Fields}
\label{sec_explicit}
For this case, $f$ and $\mathcal{ST}$ are parameterized by an explicit representation, like a feature grid.
Considering the term $\pderiv{f(p; \bTheta_f)}{p}$, $f$ can be written as an interpolation function.
The oscillating gradient $\pderiv{f(p; \bTheta_f)}{p} $ during interpolation across grid corners will severely preclude the optimization process, leading to slow convergence and local minima.

\subsubsection{Implicit Fields}
Generally, implicit fields provide smooth and global gradients as all points are querying the full MLP.
However, to encode high frequency information in an MLP, a positional encoding is usually applied to $z_i$ before feeding into MLP.
Thus, $f$ will be a composition of $f(p) = f' \circ \gamma (p)$,
$\gamma_k(p) = \big[ \cos(2^{k}\pi p),
\sin(2^{k}\pi p) \big]$. As shown in \citet{lin2021barf}, the positional encoding will amplify the gradient exponentially, which leads to unstable training.

\subsection{Gradient in Point-based Rendering}

In contrast to volume rendering, point-based rendering works with discrete representations directly, which requires a primitive organization $\mathcal{O}$ (with optional gauge transformation $\mathcal{T}$).
For clarity, we denote the joint process of primitive organization and gauge transformation as $\mathcal{OT}(p; \bTheta_{ot})$, where $\bTheta_{ot}$ and $p$ are the primitive organization \& gauge transformation parameters and a certain point.
With an initialized discrete point $p$, the process to yield a new discrete point $p'$ can be written as $\mathcal{OT}(p; \bTheta_{ot})=p'$.
To this end, the color contribution from this point can be formulated as 
\begin{equation}
    G(p', r) = G( \mathcal{OT}(p; \bTheta_{ot}); r),
\end{equation}
where $G$ is the splat-based rendering function, $r$ is the target ray.
Compared with the case of volume rendering in eq.~(\ref{eq_vr}), there is no representation function $f$ in point-based rendering, which simplifies the gradient analysis.
Then the gradient $\J$ of the color contribution with respect to $\bTheta_{ot}$ can be derived as:
\begin{equation}
\label{eq_splat_gradient}
    \J = \pderiv{G(p', r )}{\bTheta_{ot} } 
    = \pderiv{G(p', r)}{p'} \pderiv{\mathcal{OT}(p; \bTheta_{ot}) }{\bTheta_{ot}}.
\end{equation}
Notably, the gradient $\pderiv{G(p', r )}{\bTheta_{ot} }$ has been carefully handled in \cite{yifan2019differentiable,kerbl20233d} to achieve stable optimization.
For the term $\pderiv{\mathcal{OT}(p; \bTheta_{ot}) }{\bTheta_{ot}}$, its optimization (or gradient characteristic) depends on the parameterization types of $\mathcal{OT}$.

The cases of grid-based and MLP-based parameterization have been analyzed in Sec. \ref{sec_explicit}.
Notably, the parameterization of $\mathcal{OT}$ can actually be discarded for point-based rendering, which means the parameters of $\mathcal{OT}$ can be simply saved as additional properties of discrete points.
For the case without parameterization, the gradient in eq.~(\ref{eq_splat_gradient}) is stable as $\pderiv{\mathcal{OT}(p; \bTheta_{ot})} {\bTheta_{ot}}$ is a constant with respect to $\bTheta_{ot}$.

\begin{figure}[ht]
    \includegraphics[width=1.0\linewidth]{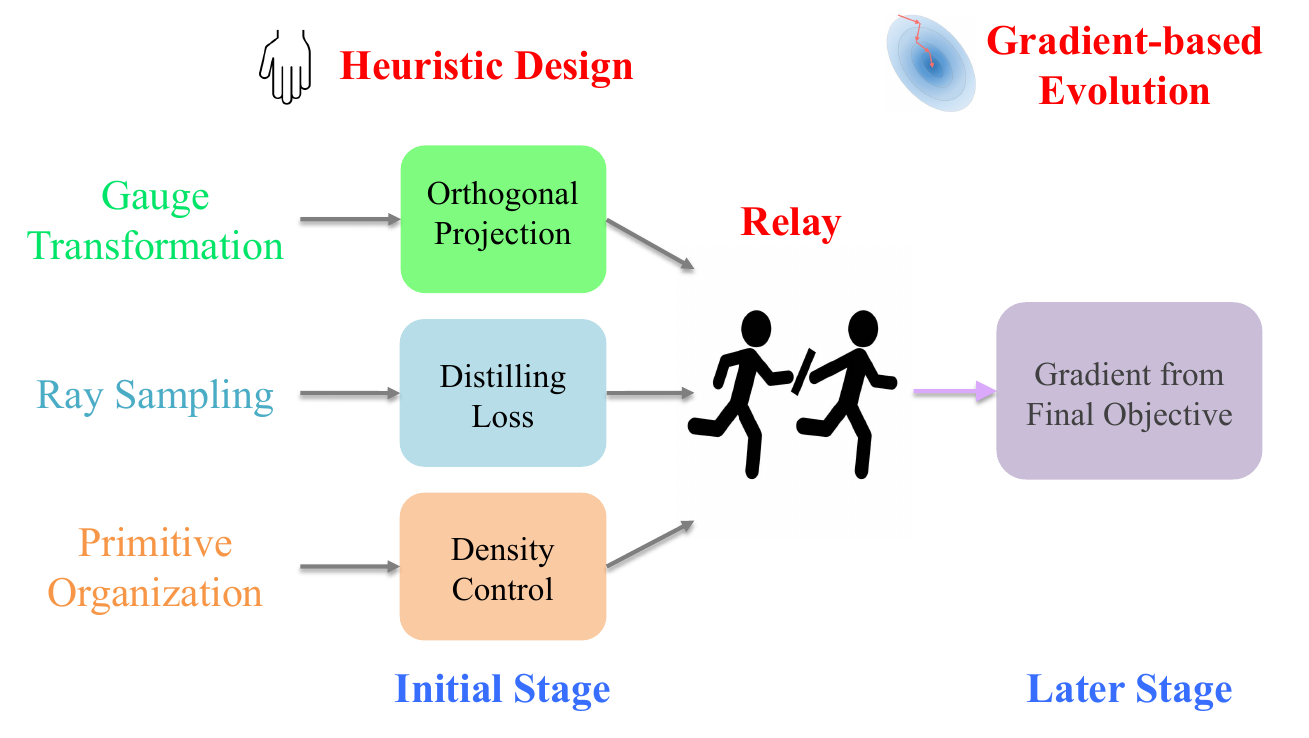}
    \caption{
        An illustration of relay learning mechanism. At the initial stage, the optimization is performed with heuristically designed elements, e.g., orthogonal projection for transformation, distilling loss for learning ray sampling, density control \cite{kerbl20233d} for primitive organization.
        After certain iterations, the optimization is relayed to the gradient-based elements evolution.
        }
    \label{fig_relay}
\end{figure}

\subsection{Relay Learning Mechanism}
Overall, we observe the consistent gradient problem (\eg, fluctuation, large value), which are especially severe at the initial training stage, and will became smoother as the training goes.
Motivated by our derivation and observation, we introduce a relay learning mechanism to facilitate the training process and avoid local minima when optimizing the rendering elements as shown in Fig. \ref{fig_relay}.
Specifically, heuristically designed elements are employed at the initial training stage to achieve stable training and approximate the optimal solution, followed by the evolutive elements for accurate optimization towards the optimal solution.
This mechanism ensures that the optimization will not suffer from the gradient problem at initial stage, and can effectively utilize the smooth gradient at later stages.
By default, we perform the optimization relay around the first 10\% steps.

\hspace{5pt}
\section{Experimental Evaluation}

We evaluate the effectiveness of ERM by replacing the heuristic design in existing rendering models with our evolutive elements.

\begin{figure*}[ht]
    \includegraphics[width=1.0\linewidth]{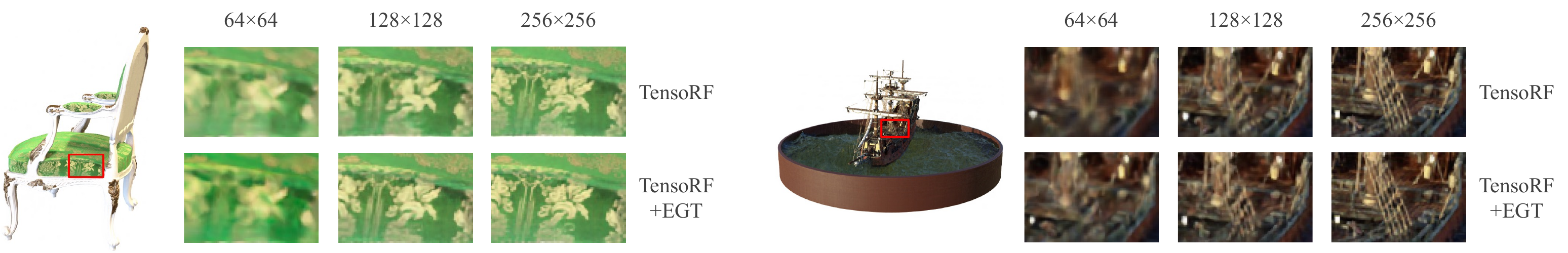}
    \caption{
    Comparison between renderings with the inclusion of EGT under different plane size.
        }
    \label{fig_gauge_vis}
\end{figure*}

\subsection{Evolutive Gauge Transformation}

To validate the effectiveness of our evolutive gauge transformation (EGT), we replace the orthogonal projection in TensoRF, KPlanes, and EG3D with our learnable mapping, to perform static scene modeling, dynamic modeling, and generative modeling, respectively.

\renewcommand\arraystretch{1.1}
\begin{table}[t]
\renewcommand\tabcolsep{7.25pt}
\centering 
\begin{tabular}{lcccc} 
\hline
Models & Setting
& Time & PSNR$\uparrow$ & SSIM$\uparrow$ 
\\
\hline

\multicolumn{5}{c}{\textbf{Static Modeling on Synthetic NeRF}} \\

PlenOctrees &  N/A     & $\sim$15 hr & 31.71  & 0.958   \\
Plenoxels &  N/A     & 11.4 min & 31.71  & 0.958 \\
DVGO &  N/A     & 15.0 min & 31.95  & 0.957   \\
Mip-NeRF & N/A    & 2.89 hr &  33.06  & 0.960  \\

TensoRF & 256 $\times$ 256      & 12.5 min & 33.01   & 0.963   \\
\rowcolor{LightCyan}  
TensoRF+EGT & 256 $\times$ 256      & 13.1 min  & \textbf{33.38}  & \textbf{0.964}    \\

\hline

\multicolumn{5}{c}{\textbf{Dynamic Modeling on D-NeRF}} \\

KPlanes & 256 $\times$ 256       & 36.2 min & 31.03    &  0.946    \\
\rowcolor{LightCyan}  
KPlanes+EGT & 256 $\times$ 256   & 36.6 min & \textbf{31.31}   & \textbf{0.947}  \\

\hline

Models& Setting & Time & \multicolumn{2}{c}{FID$\downarrow$}  \\
\hline

\multicolumn{5}{c}{\textbf{Generative Modeling on FFHQ}} 
\\

EG3D & 256 $\times$ 256       & 46 hr & \multicolumn{2}{c}{7.051}    \\
\rowcolor{LightCyan}  
EG3D+EGT &  256 $\times$ 256   & 54 hr & \multicolumn{2}{c}{\textbf{6.546}}  \\

\hline

\end{tabular}

\caption{
Evaluation of our evolutive gauge transformation at various tasks including static scene modeling, dynamic modeling, and generative modeling.
TensoRF, KPlanes, and EG3D serve as the base model respectively.
The setting indicates the size of feature planes. The plane size for modeling gauge transformation is kept the same as the base model by default.
}
\label{tab_gauge_compare}
\end{table}

\begin{table*}[t]
\small 
\renewcommand\tabcolsep{4.2pt}
\centering 
\begin{subtable}[c]{0.175\textwidth}

\begin{tabular}{lccccc} 
\hline
& \multicolumn{2}{c}{\textbf{Sampling}} & 
\multicolumn{3}{c}{\textbf{Synthetic NeRF}}
\\
\cmidrule(lr){2-3} \cmidrule(lr){4-6}
\multirow{-2}{*}{\textbf{Models}} & Coarse & Final
& Time$\downarrow$ & PSNR$\uparrow$ & SSIM$\uparrow$ 
\\\hline

\textbf{NeRF} & 32 & 64      & 8.81 hr & 28.78   & 0.933   \\
\rowcolor{LightCyan}  
\textbf{NeRF+ERS} & 32 & 64      & \textbf{8.43} hr  & \textbf{30.29}  & \textbf{0.941}    \\

\hline

\textbf{NeRF} & 64 & 64       & 10.5 hr & 29.76     &  0.941    \\ 
\rowcolor{LightCyan}  
\textbf{NeRF+ERS} & 64 & 64   & \textbf{10.1} hr & \textbf{30.90}   & \textbf{0.946}  \\

\hline

\textbf{NeRF} & 96 & 64       & 12.0 hr & 30.79    &  0.946    \\
\rowcolor{LightCyan}  
\textbf{NeRF+ERS} & 96 & 64   & \textbf{11.6} hr & \textbf{31.21}   & \textbf{0.947}  \\

\hline

\end{tabular}
\end{subtable}
%
%
%
%
\hfill
\begin{subtable}[c]{0.585\textwidth}
\begin{tabular}{lcccccccc} 
\hline
& \multicolumn{2}{c}{\textbf{Sampling}} 
& \multicolumn{3}{c}{\textbf{Synthetic NeRF}} 
& \multicolumn{3}{c}{\textbf{D-NeRF}}
\\
\cmidrule(lr){2-3}\cmidrule(lr){4-6}\cmidrule(lr){7-9}
\multirow{-2}{*}{\textbf{Models}} & Coarse & Final
& Time$\downarrow$ & PSNR$\uparrow$ & SSIM$\uparrow$
& Time$\downarrow$ & PSNR$\uparrow$ & SSIM$\uparrow$ 
\\\hline

\textbf{KPlanes} & 32 & 48      & 27 min & 27.93   & 0.950  &       
  41 min  & 29.13   & 0.962  \\

\rowcolor{LightCyan}
\textbf{KPlanes+ERS} & 32 & 48       & \textbf{26} min & \textbf{30.70}  & \textbf{0.957}  & 39 min  & \textbf{30.21}   & 0.964   \\

\hline

\textbf{KPlanes} & 64 & 48        & 28 min    & 30.02   & 0.958   &  39 min & 30.49 & 0.967  \\
\rowcolor{LightCyan} 
\textbf{KPlanes+ERS} & 64 & 48    & 28 min   & \textbf{31.85}  & \textbf{0.961}   & 41 min  & \textbf{30.95} & 0.969  \\

\hline

\textbf{KPlanes} & 96 & 48        & 30 min     &  31.84  & \textbf{0.961}   & 42 min  & 31.03 & 0.969   \\
\rowcolor{LightCyan}  
\textbf{KPlanes+ERS} & 96 & 48    & 30 min   & \textbf{32.29}  & \textbf{0.962}   & 42 min & \textbf{31.29}  & \textbf{0.970}  \\
\hline

\end{tabular}
\end{subtable}

\caption{
The rendering performance by integrating evolutive ray sampling.
`Coarse' and `Final' denote the number of points for sampling fields and radiance fields.
}
\label{tab_sampling}
\end{table*}

\subsubsection{Static Scene Modeling}
We first evaluate our evolutive gauge transformation on the static scenes from the Synthetic NeRF dataset \cite{mildenhall2020nerf}.
Here, we use TensoRF \cite{chen2022tensorf} as the baseline model, with a plane size of $256\times 256$ and a plane dimension of 64.
For the gauge transformation, we adopt the same model structure and plane size as the base model.
The model is trained with a pre-defined orthogonal projection for the first 3000 steps, and subsequently the optimization transitions into using our proposed evolutive gauge transformation (EGT) to learn a flexible mapping.
Intuitively, with the orthogonal mapping as the initizliation, we only learn a residual transformation term as illustrated in Fig. \ref{fig_gauge_compare}.

As shown in Table \ref{tab_gauge_compare}, the inclusion of EGT leads to a clear gain in the rendering quality of TensoRF, while only slightly reducing the training speed. Notably, the performance gain will be more distinct with decreasing feature plane sizes. We conjecture that the gradient oscillation around the grid corner will be mitigated with a small plane size, which leads to more stable optimization as analyzed in Sec. \ref{sec_explicit}.

\subsubsection{Dynamic Scene Modeling.}
For dynamic scene modeling on the D-NeRF dataset \cite{pumarola2020dnerf}, we set KPlanes as the base model with a plane size of $256\times 256$ and a feature dimension of 16 \footnote{We adopt smaller feature plane size and dimension as we find the original KPlanes setting for D-NeRF is redundant.}.
The EGT employs the same model structure and plane size as the base model.
As shown in Table \ref{tab_gauge_compare}, consistent performance improvements can be observed with the integration of EGT.

\subsubsection{Generative Scene Modeling.}
EG3D \cite{chan2022efficient} serves as the base model for 3D generative modeling.
Specifically, the Triplane structure and plane size ($256\times 256$) in EG3D is also adopted for the gauge transformation.
The Triplane for gauge transformation is generated from the latent code with the generator in StyleGAN2.
The training is performed with predefined orthogonal projection at the initial stage (10\% of total iteration).

As shown in Table \ref{tab_gauge_compare}, the FID of the generated images can be improved by 0.505, while the training time will be increased by 8 hours as generating additional Triplanes with StyleGAN2 for the gauge transformation is a cumbersome process.

\begin{figure*}[ht]
    \includegraphics[width=1.0\linewidth]{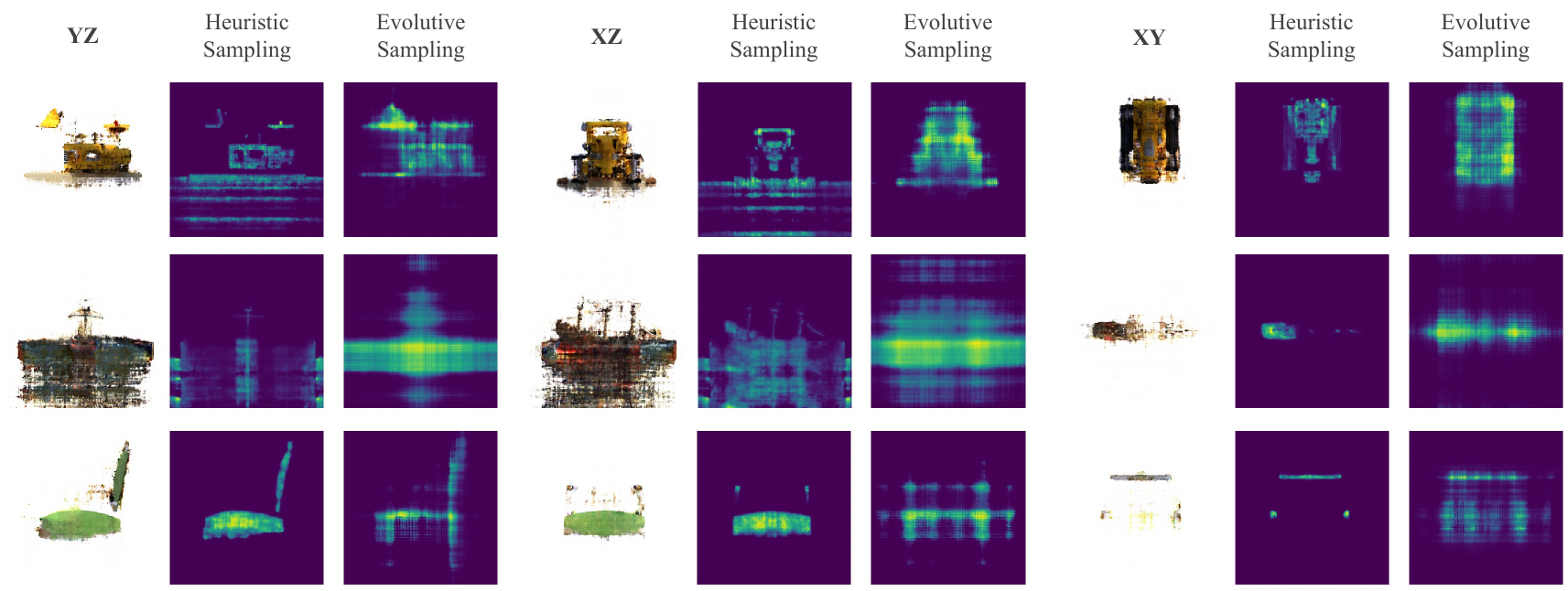}
    \caption{
        The illustration of learned sampling fields in heuristic design and our evolutive method. 
        We visualize three 2D orthogonal cross sections (\ie, YZ, XZ, and XY) of the sampling fields.
        The scenes include Lego, Ship, and Chair from the Synthetic NeRF dataset.
        }
    \label{fig_sampling_vis}
\end{figure*}

\subsection{Evolutive Ray Sampling}
We evaluate the performance of ERS on static scene modeling and dynamic scene modeling.

\subsubsection{Static Scene Sampling}
We perform experiments on the Synthetic NeRF dataset with NeRF and KPlanes as the base models.
The plane size in KPlanes is set as $256\time 256$ with a feature dimension of 32. The sampling field adopts a plane size of $64\times 64$ with a feature dimension of 8.
Specifically, both NeRF and KPlanes are equipped with sampling fields to perform coarse-to-fine sampling.
To train the sampling fields, NeRF and KPlanes adopt a recontruction loss and distillation loss, respectively.
NeRF+ERS and KPlanes+ERS remove these reconstruction losses and the distillation loss, as they can directly train the sampling fields by propagating gradient from the final training loss through the sampling process. 

As shown in Table \ref{tab_sampling}, the rendering quality and training time are improved consistently with the inclusion of ERS.
We also ablate the effect of different number of coarse sampling ppoints, and observe that the ERS is more robust to the number of sampling points compared with the previous heuristic design.
Notably, the NeRF training speed is also slightly improved as rendering operations in the reconstruction loss are reduced.

We also visualize and compare the learned sampling fields in Fig. \ref{fig_sampling_vis}.
Specifically, we take three 2D orthogonal cross-sections (\ie, YZ, XZ, and XY section) of the volume, which are uniformly sampled to query the sampling fields to get color and density. 
As shown in Fig. \ref{fig_sampling_vis}, the learned sampling fields with heuristic design are not well aligned with the scene geometry \& surface as it is trained with reconstruction loss.
As the comparison, the sampling fields learned with ERS tends to be smoothly distributed around scene surface.
We conjecture this geometry slack of sampling fields is more beneficial for flexible volume rendering as there is no harsh geometry constraint.

\subsubsection{Dynamic Scene Sampling}
We validate the effectiveness of ERS with KPlanes as the base model on D-NeRF dataset.
The KPlanes and sampling fields settings are similar to case of static scene modeling, just including an additional time dimension of size 50 for KPlanes and 25 for sampling fields.
As shown in Table \ref{tab_sampling}, the performance gain is also consistent with the inclusion of ERS.

\begin{table*}[t]
\small
\renewcommand\tabcolsep{2.0pt}

\begin{subtable}[c]{0.55\textwidth}

\begin{tabular}{lcccccccccccc} 
\hline
& \multicolumn{6}{c}{\textbf{Mip-NeRF360}} & 
\multicolumn{6}{c}{\textbf{Tanks\&Temples}}
\\
\cmidrule(rl){2-7} \cmidrule(lr){8-13}
\multirow{-2}{*}{\textbf{Method}} & SSIM$\uparrow$ &PSNR$\uparrow$ &  LPIPS$\downarrow$ & Train & FPS & Mem
& SSIM$\uparrow$ & PSNR$\uparrow$ &  LPIPS$\downarrow$ & Train & FPS & Mem 
\\\hline

\textbf{Plenoxels}     &0.626 &23.08 &0.463 &25m49s &6.79 &2.1GB &0.719 &21.08 &0.379 &25m5s &13.0 &2.3GB \\
\textbf{INGP-Base}      & 0.671 &25.30 &0.371 &5m37s &11.7 &13MB &0.723 &21.72 &0.330 &5m26s &17.1 &13MB\\
\textbf{INGP-Big}      &0.699 &25.59 &0.331 &7m30s &9.43 &48MB &0.745 &21.92 &0.305 &6m59s &14.4 &48MB\\
\textbf{M-NeRF360}      & 0.792 &\textbf{27.69}& 0.237& 48h &0.06 &\textbf{8.6MB} &0.759 &22.22 &0.257 &48h &0.14 &\textbf{8.6MB}\\
\rowcolor{gray} \textbf{3D-GS}      & 0.815 & 27.21 & 0.214 & \textbf{41m33s} & 134 & 734MB & 0.841 & 23.14 & \textbf{0.183} & \textbf{26m54s} & 154 &  411MB\\
\rowcolor{orange} \textbf{3D-GS+EPO}     & \textbf{0.838} & 27.45  & \textbf{0.208} &47m24s&\textbf{140}&691MB  &\textbf{0.853}&\textbf{24.10}&0.193&35m17s&\textbf{161}&380MB\\

\hline

\end{tabular}
\end{subtable}
\hfill
\begin{subtable}[c]{0.275\textwidth}
\begin{tabular}{lccc} 
\hline
& \multicolumn{3}{c}{\textbf{Synthetic NeRF}}
\\
\cmidrule(lr){2-4}
\multirow{-2}{*}{\textbf{Method}} 
&SSIM$\uparrow$ &PSNR$\uparrow$ &  LPIPS$\downarrow$  
\\\hline
\textbf{Plenoxels}        &0.958 & 31.71& 0.049  \\
\textbf{Mip-NeRF}      & 0.960& 33.06& 0.043  \\
\hline
\rowcolor{gray} \textbf{P-NeRF}      & 0.967 &30.71& 0.081  \\
\rowcolor{orange} \textbf{P-NeRF+EPO}      & \textbf{0.969} &\textbf{31.53}& \textbf{0.077}  \\
\hline
\rowcolor{gray} \textbf{3D-GS}      & 0.968 &33.32    & 0.040 \\
\rowcolor{orange} \textbf{3D-GS+EPO}      &  \textbf{0.973}   &  \textbf{33.95}   &  \textbf{0.037}   \\
\hline

\end{tabular}
\end{subtable}

\caption{
EPO demonstrates superior performance on task of single scene radiance field modeling in both 3D Gaussian Splatting (3D-GS) framework and point-NeRF (P-NeRF) framework, we test on real-world mipnerf360~\cite{barron2022mip}, Tanks\&Temples~\cite{knapitsch2017tanks} scenes as well as synthetic scenes from Synthetic-NeRF \cite{mildenhall2020nerf} dataset.
}
\label{gs-singlescene}
\end{table*}

\begin{figure*}
    \centering \includegraphics[width=0.98\linewidth]{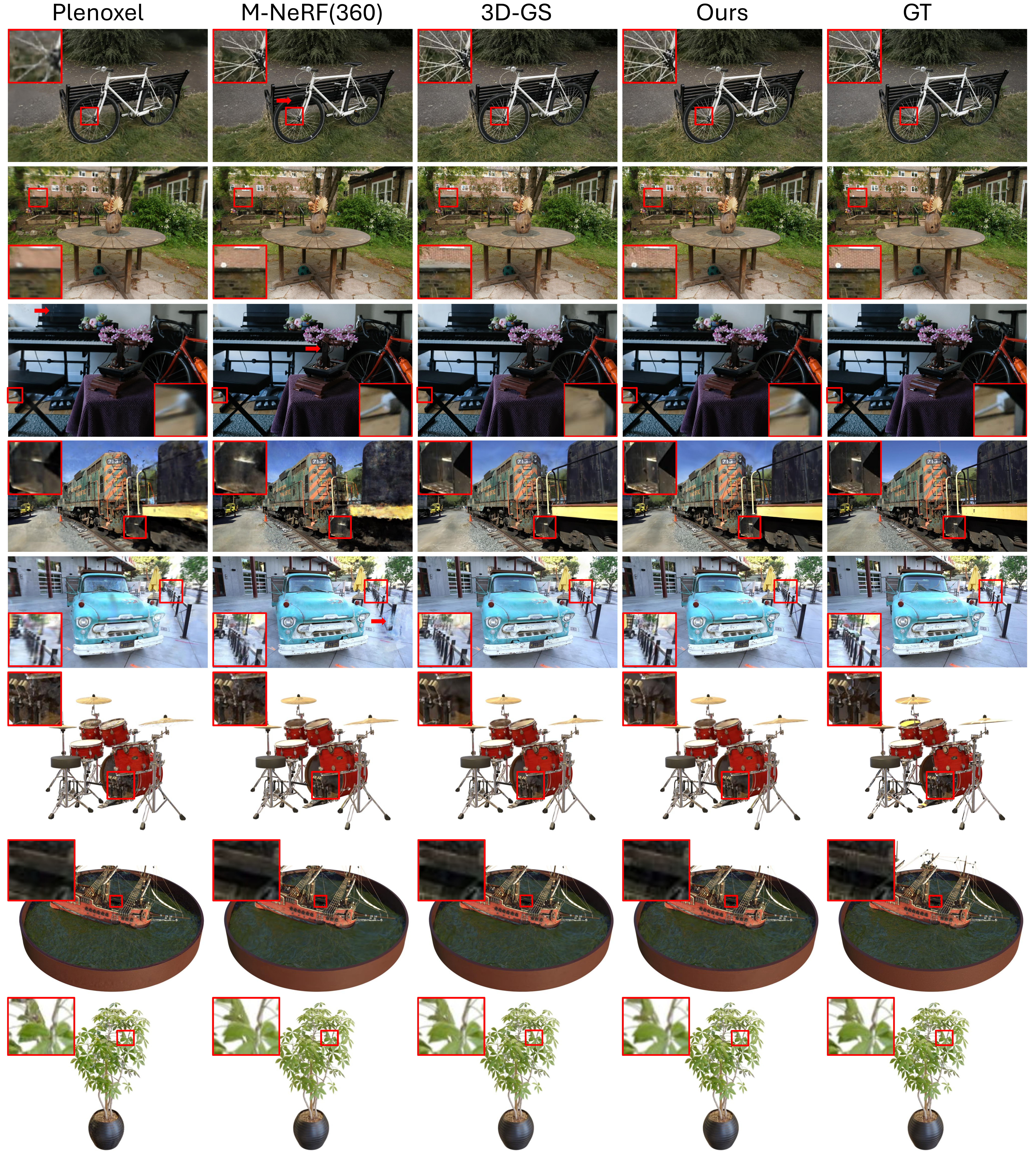}
    \caption{We show qualitative comparisons of our (Evolutive 3D Gaussian Splatting) to previous methods  and the corresponding ground truth images from held-out test views. The scenes are, from the top down: Bicycle, Garden, Bonsai from the Mip-NeRF360 dataset; Train and Truck from Tanks\&Temples; Drums, Ship, Ficus from NeRF Synthetic dataset. Differences in quality highlighted by arrows/insets.}
\label{fig:evolutiongs-vis}
\end{figure*}

\subsection{Evolutive Primitive Organization}
To evaluate the effectiveness of evolutive primitive organization (EPO), we test our component on both 3D Gaussian Splatting (3D-GS) framework and point-NeRF (P-NeRF) framework on the task of single scene radiance field rendering.

\subsubsection{Evolutive 3D Gaussian Splatting} 

For 3D-GS, we regard each Gaussian as scene primitive and both the growing and splitting process will be learned during the whole optimization process. The original 3D-GS method will directly clone the Gaussians during the growing operation and does not have differentiable sampling and splitting operations. In our learned growing process, we will learn a position term $\delta\mu$ that defines the posituion of the newly split Gaussians. 
To do that, we apply spherical distribution growth to learn growth directions of new Gaussians. We implement the grown probability $Q$ as an attribute of existing Gaussians and directly optimize it throughout whole optimization process. When using Gaussians to learn radiance field, the newly-grown Gaussians should not be too far away from old Gaussians. Thus to learn a reasonable growth distance, instead of applying radial distribution growth, we can directly learn the distance by using standard deviation of original Gaussians as a constraint. More specifically, we set $\|\delta\mu\|=v*(1/1+exp(-s))$, where $s$ is learnable parameter in our implementation and $v$ is two times maximum standard deviation of original Gaussians. The other properties (scales, sh coefficients etc.) of newly-grown Gaussians are copied from original ones.

In addition to the learned growth, we also propose a learned splitting operation in our model. As mentioned in Sec.~\ref{sec:EPO}, we learn a split mean shift term that decide the position of newly-split Gaussians. The split mean shift is formulated as $\delta\mu_k=R(\sigma_k*(1/1+exp(-s')))$, where $\sigma_k$ and $R$ are the standard deviation and rotation matrix of original Gaussians. $s'$ is the learned parameter that control the length of the split shift. In addition to that, we also learn a scaling factor $\phi=1.2*(1/1+exp(-v))+1$ for each split Gaussian, where $v$ is the learned scalar parameter, and the newly split Gaussian scale will be divided by scaling factor $\phi$. Similar to that in 3D-GS, our differentiable growing and splitting operations focus on not well reconstructed region with large view-space positional gradients.

We follow the same training and evaluation setup as in 3D-GS. Similiar to that in 3D-GS, we initialize the position of 3D Gaussians using SFM points except for NeRF synthetic scenes where the Gaussian positions are randomly initialized. We set the number of potential directions $N$ to be 128 and do the  growing and splitting operation every 100 training iterations. Each scene is optimized for 30k iterations.

\textbf{Results}
We test our model on both real-world scenes from previously published datasets, including full set of scenes from Mip-NeRF360 dataset \cite{barron2022mip}, eight scenes from LLFF dataset \cite{mildenhall2019local}, two scenes from Tanks\&Temples dataset \cite{knapitsch2017tanks}, and synthetic scenes from the synthetic Blender dataset \cite{mildenhall2020nerf}. Those scenes have various capture styles, and cover both
bounded indoor scenes and large unbounded outdoor environments.

\textit{Real-World Scenes} We compare our method against several state-of-the-art techniques including Mip-NeRF360, 3D-GS as well as recent fast NeRF methods: InstantNGP 
and Plenoxels. We report results for a
basic configuration of InstantNGP (Base) as well as a slightly larger network suggested by the authors (Big). We take every 8th images for test set and others for train set and compare with the standard PSNR, L-PIPS, and SSIM metrics, please see Table \ref{gs-singlescene} and Table \ref{tab_po_ablation}.

In contrast to Mip-NeRF 360, our model attains comparable results on the Mip-NeRF360 dataset and significantly outperforms it on the Tanks \& Temples dataset. Furthermore, our model exhibits markedly faster training and inference speeds. Notably, compared to the original 3D-GS, our method achieves superior performance with, on average, fewer Gaussians per scene. This leads to reduced memory requirements and accelerated rendering speeds. This improvement is attributed to the efficacy of the evolution strategy, which enables more effective growth and splitting of Gaussians. We also show qualitative results of this comparison on
test view in Fig.~\ref{fig:evolutiongs-vis}. Compared with previous methods, our model tends to preserve more visual detail from far away (Scene GARDEN, TRUCK and TRAIN) and recover some delicate thin structures (Scene BICYCLE, BONSAI), while original 3D-GS and Mip-NeRF360 may fail at those circumstances.

\begin{table}[t]
\centering 
\renewcommand\tabcolsep{4.85pt}
\centering 

\begin{tabular}{lccc} 
\hline
& \multicolumn{3}{c}{\textbf{LLFF}}
\\
\cmidrule(lr){2-4}
\multirow{-2}{*}{\textbf{Method}} 
& PSNR$\uparrow$ & SSIM$\uparrow$ & LPIPS$\downarrow$ 
\\\hline

\textbf{3D-GS}      & 25.82 & 0.821   & 0.200   \\
\textbf{3D-GS+LGD}      & 24.37 & 0.796   & 0.214   \\
\textbf{3D-GS+LSDG (Soft)}      & 26.35 & 0.838   & 0.195   \\
\textbf{3D-GS+LSDG (Repara)}      & 26.76 & 0.881   & 0.186   \\
\textbf{3D-GS+LSDG (Repara)+LSO}      & \textbf{27.01} & \textbf{0.890}   & \textbf{0.184}   \\

\hline

\end{tabular}

\caption{
Ablation study on different components of Evolutive 3D Gaussians Splatting on LLFF \cite{mildenhall2019local} dataset.
}
\label{tab_po_ablation}
\end{table}

\textit{Synthetic Blender Scenes} In addition to realistic scenes, we also
evaluate our approach on the synthetic NeRF dataset, results in Fig~\ref{gs-singlescene}. Even though our approach starts training from 100K uniformly random Gaussians inside a volume that encloses the scene bounds, our approach can quickly converge to reasonable Gaussians, with better performance than all previous state-of-the-art methods. Similarly to the case in real-world scene, the Gaussians in our model grows and splits in a more efficient way, resulting in modeling the radiance field with fewer Gaussians which achieving better performance compared to 3D-GS.

\textbf{Ablations}
In this part, we evaluate different components of our evolutive Gaussians design on LLFF dataset, including the learned spherical distribution  growth (LSDG) and learned splitting operation (LSO), results shown in Table~\ref{tab_po_ablation}. For LSDG design, we compare against the straightforward method of directly learning  growth direction (LGD). In our learned spherical distribution growth design, we choose to grow along the direction of maximum probability and propose the reparameterization (Repara-) strategy for optimization as shown in Alg. ~\ref{alg1}. To validate this proposal, we also test a soft variant to learn spherical distribution growth (Soft-), where grow direction is decided as weighted sum of all possible directions $d=\sum_{i=1}^{N}p_{i}d_{i}$. 

When adopting the naive way of directly learning the growth direction, we find that the performance is even worse than baseline model. We speculate that it is because directly learning the growth direction has over-flexibility which makes the model to be vulnerable to local minimum. Compared to Soft version of learning spherical distribution growth, our reparameterization design adopts a more reasonable way of choosing growth direction, resulting in a better performance. The inclusion of learning splitting operation helps to further boost the performance. Our complete model is able to outperform original 3D-GS on PSNR by a significant margin of 1.2 dB, demonstrating the effectiveness of our evolutive primitive organization design.

\begin{figure}[ht]
    \includegraphics[width=1.0\linewidth]{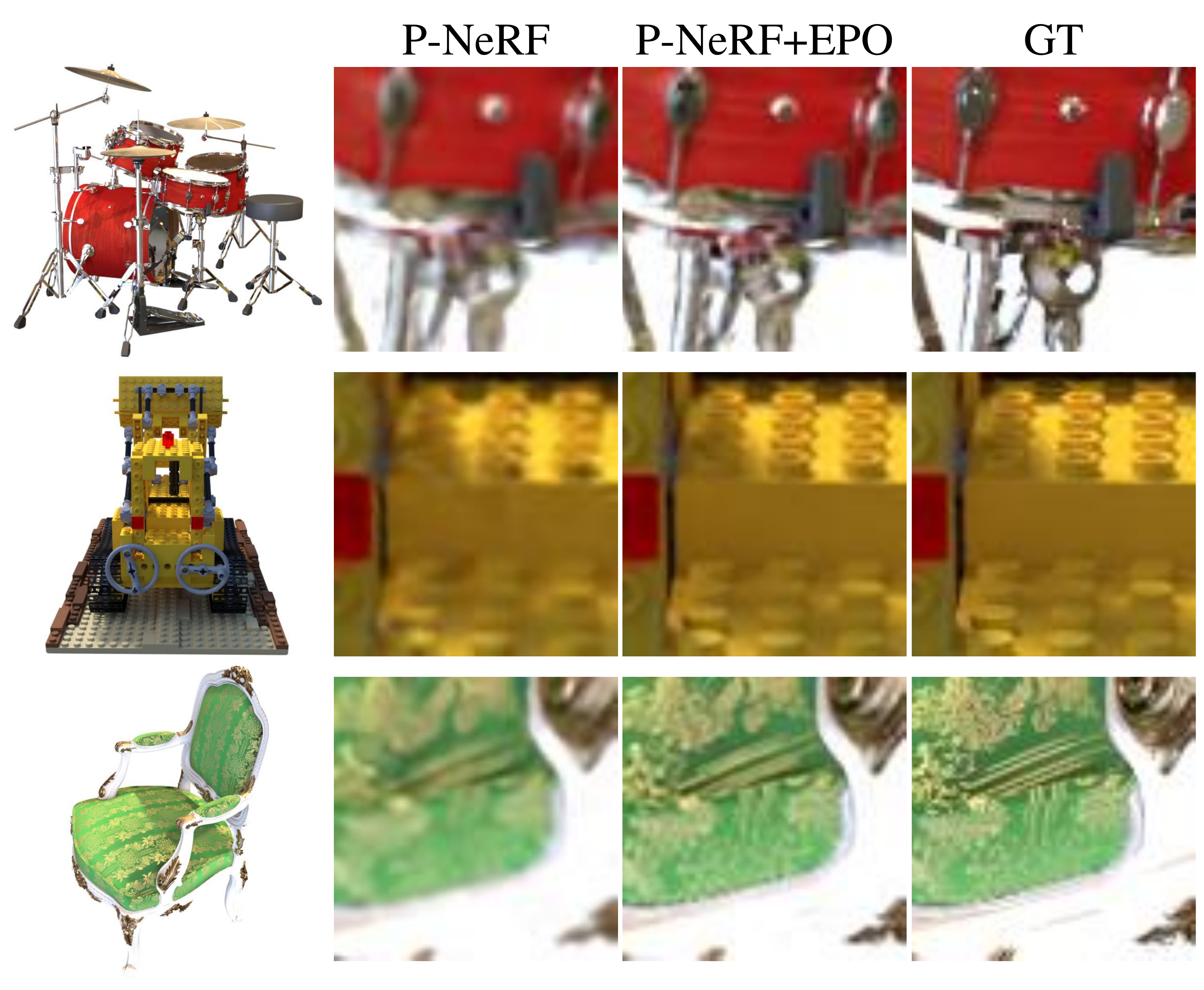}
    \caption{
        We show qualitative comparisons between P-NeRF~\cite{xu2022point} with our method that combines P-NeRF with evolutive primitive organization (EPO), as well as the corresponding ground truth images from held-out test views.
        }
    \label{fig:pnerf-vis}
\end{figure}

\subsubsection{Evolutive Neural Point} In addition to evolutive 3D Gaussian Splatting, we also test Evolutive Primitive Organization in the framework of Point-NeRF (P-NeRF). In P-NeRF, the whole scene is composed of neural points feature that encode the radiance field. Instead of using splatting as in 3D-GS, P-NeRF adopts volume rendering mechanism where sampled points along the ray will query feature from neighbouring neural points, which will then be decoded into density and rgb color space. In P-NeRF, they adopts hand-designed growing and pruning operation to avoid holes and outliers in initial points. These operations have similar issue as in 3D-GS, that the operations are no-differentiable and may misalign with the final objective.

To alleviate this issue, we regard each neural point as scene primitive and learn the growing operation in per-scene optimization process. Particularly, the position of new neural points will be that of old neural points plus a learnable growth term $\delta\mu$. We apply spherical distribution growth to learn growth directions of new neural points. Similar to that in 3D-GS, the growth distance will be learned directly as $\|\delta\mu\|=v*(1/1+exp(-s))$, where $s$ is learnable parameter in our implementation and $v$ is two times initial voxel grid size in P-NeRF.

We follow the same per-scene optimization setup as in P-NeRF,  where we adopt a loss function that combines the rendering and the sparsity loss. We do per-scene training for 20k iterations and
perform point growing and pruning every 1K iterations.
we evaluate our approach on the synthetic NeRF dataset, results in Table.~\ref{gs-singlescene}. Compared to P-NeRF baseline method, the adoption of evolution neural points helps to improve performance on all metric, especially on PSNR and LPIPS by a good margin. This proves that our Evolutive Primitive Organization is effective in both volume rendering and splatting mechanism. Qualitative result is shown in Fig.~\ref{fig:pnerf-vis}.

\renewcommand\arraystretch{1.1}
\begin{table}[t]
\renewcommand\tabcolsep{12pt}
\centering 
\begin{tabular}{lccc} 
\hline
 & \multicolumn{3}{c}{\textbf{UV Mapping on DTU}} \\

\cmidrule(lr){2-4}
\multirow{-2}{*}{\textbf{Models}}  & Regularization
& PSNR$\uparrow$ & SSIM$\uparrow$ 
\\\hline

\hline
NeuTex & \textcolor{green}{\checkmark}        & 28.02    &  0.891    \\
NGF & \textcolor{green}{\checkmark}        & 27.74    &  0.887    \\
\rowcolor{LightCyan}   Ours &  \textcolor{red}{$\times$}    & \textbf{29.41}   & \textbf{0.907}  \\
\hline

\end{tabular}

\caption{
Applications in UV mapping and editing.
}
\label{tab_gauge_uv}
\end{table}

\begin{figure}[ht]
    \includegraphics[width=1.0\linewidth]{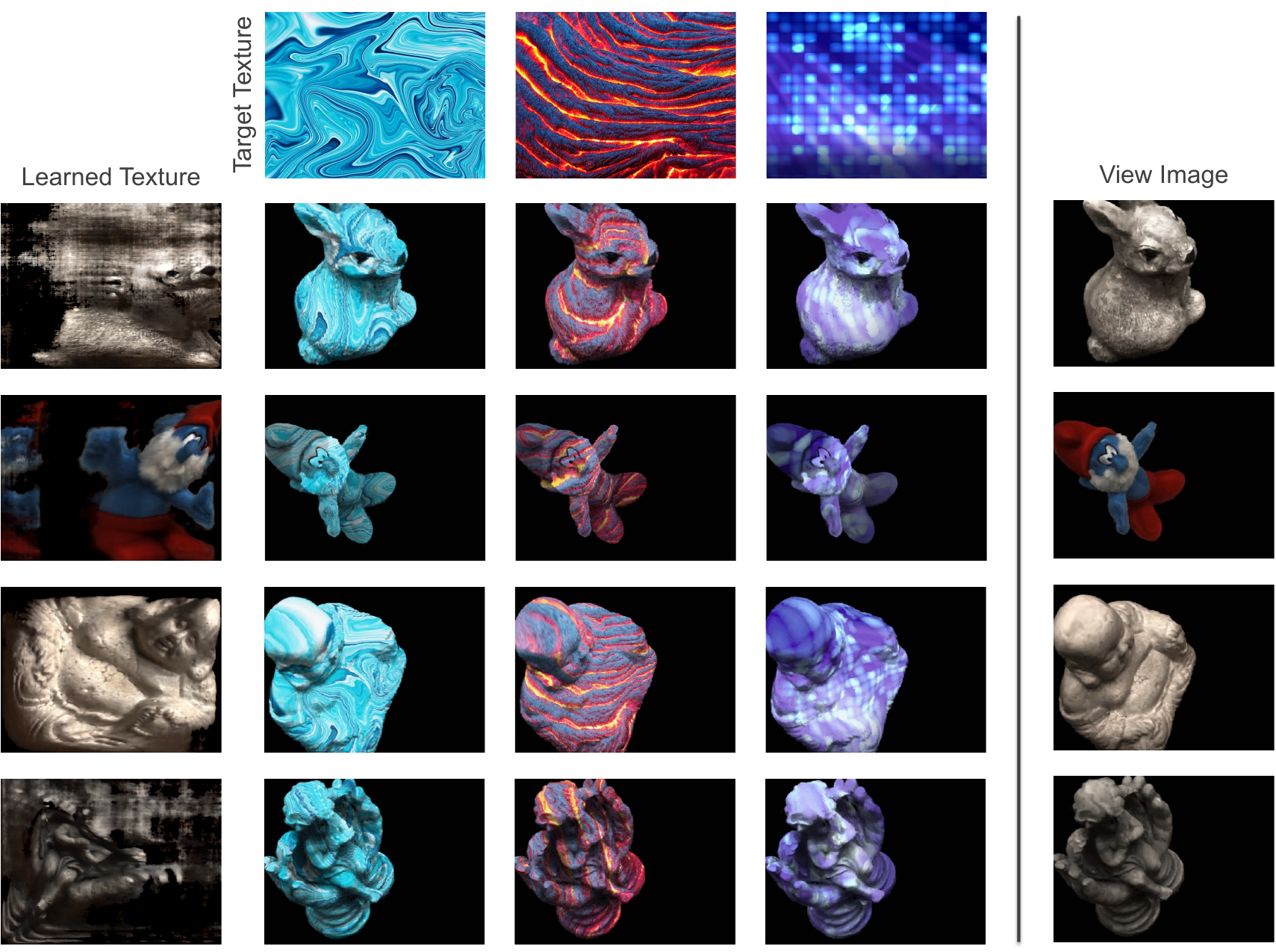}
    \caption{
        UV editing results.
        }
    \label{fig_gauge_uv}
\end{figure}

\begin{figure*}
    \centering \includegraphics[width=0.95\linewidth]{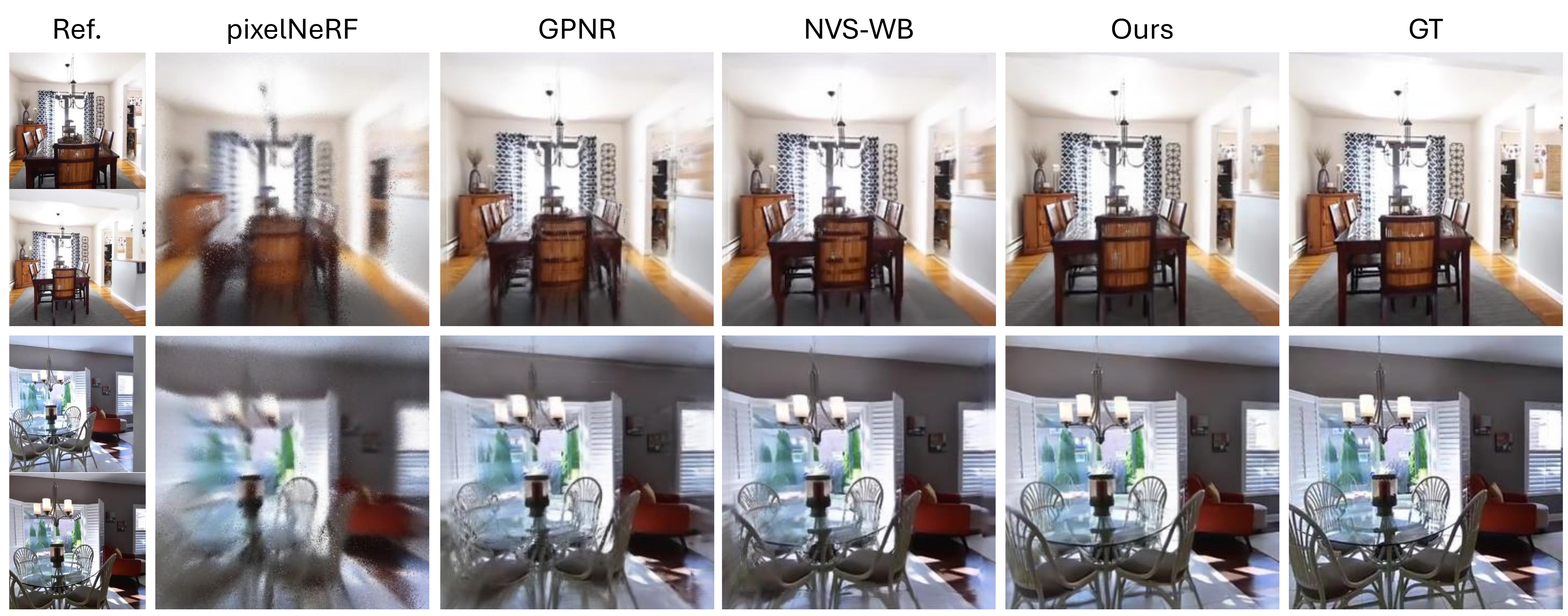}
    \caption{We show qualitative comparisons of ours to previous methods and the corresponding reference images from RealworldEstate10k dataset.}
\label{fig:generalizable-gs}
\end{figure*}

\section{Applications}
\subsection{UV Mapping}
Learning UV mapping is a non-trivial task in radiance fields.
As UV mapping aims to connect the 2D space and 3D space, our evolutive gauge transformation is a good fit to achieve it.
For this case, we aims to learn a mapping from 3D space to 2D plane, and adopt NeRF as the base model with a relay learning mechanism.
Specifically, we train a NeRF with identical mapping (\ie, without learning gauge transformation) for the first 30000 steps.
As the geometry emerges which means the gradient becomes stable, we replace the identical mapping with the learnable transformation from 3D space to 2D plane in the color branch (Note the transformation is not applied in the density branch).
After training, a UV map can be obtained by querying the radiance field with uniformly sampled points on the 2D plane.

We compare the performance of our method with NGF \cite{zhan2023general} and Neutex \cite{xiang2021neutex} as shown in Table \ref{tab_gauge_uv}. Our method outperforms previous methods on the rendering quality with UV mapping on DTU dataset \cite{aanaes2016large}. 
With the obtained UV maps, we easily edit the UV to the target texture as shown in Fig. \ref{fig_gauge_uv}.

\subsection{Generalizable Gaussian Splatting} 

3D-GS has demonstrated remarkable performance and real-time rendering through rasterization-based rendering. However, the need for retraining on each new scene limits its practical applications. One way to tackle this bottleneck is to combine the 3D-GS representation with cross-scene generalizable NeRF models, which can directly synthesize novel views of unseen scenes. To achieve that, we need to build a model that can directly predict 3D Gaussian parameters in a feed-forward manner given images of new scene. 
Thus, the gradient should be able to be back-propagated to control the grow and prune of Gaussians. Our evolutive primitive organization exactly maintains the gradient from training objective where scene primitives will be implicitly grown. It turns to be a promising optimization solution for training generalizable 3D-GS model.

More specifically, given source view images and their camera parameter, our model first use an transformer-based encoder that aggregates multi-view image features via epipolar attention~\cite{charatan2023pixelsplat} to predict pixel-wise feature. Then each feature will be passed through a decoder to directly predict the parameter of Gaussians along each ray. With known camera parameter, the direction along which to grow Gaussians has been decided. Thus we apply radial distribution growth method to predict the position of nascent Gaussians. We divide each ray into $N$ bins and learn the probability indicating the likelihood of Gaussians locating in each bin. And we decide the position of Gaussians by choosing the bin of maximum likelihood. The reparameterization  strategy ( shown in Alg. ~\ref{alg1}) will be used here for optimization.

We evaluate our model on the task of wide-baseline novel view synthesis from stereo image pairs, conducting experiments on the RealEstate10k~\cite{re10k} dataset. Following previous baseline \cite{du2023learning}, we conduct experiments on image of resolution 256x256. In multi-view image encoder, we use a DINO ~\cite{caron2021emerging} pretrained ResNet-50~\cite{he2016deep} followed by a ViT-B/8 vision transformer~\cite{dosovitskiy2020image}. Adam optimizer~\cite{kingma2014adam} is used for training.

We compare our model against three novel view-synthesis baselines, including GNPR~\cite{gpnr}, pixelNeRF~\cite{yu2021pixelnerf} and NVS-WB~\cite{du2023learning}.
GNPR uses
a vision transformer-based backbone to compute epipolar
features, and a light field-based renderer to compute pixel
colors. pixelNeRF decodes pixel-aligned feature into neural
radiance fields. NVS-WB uses a multi-view self-attention encoder and combines light
field rendering with an epipolar transformer. As shown in Table.~\ref{generalizable-gs}, our model is significantly more efficient than all 
the baselines models, inheriting the advantage of using Gaussians as scene representation. Particularly, our model is more than 100 times faster than the second best baseline model.  Meanwhile, our model is able to outperform the baselines on all metrics. These are attributed to the differentiable nature of our evolutive primitive organization module. Qualitative results are shown in Fig.~\ref{fig:generalizable-gs}.

\begin{table}[t]
\centering 
\renewcommand\tabcolsep{4.85pt}
\centering 
\begin{tabular}{lcccc} 
\hline
\textbf{Method}
& PSNR$\uparrow$ & SSIM$\uparrow$ & LPIPS$\downarrow$ & Inference Time (s) $\downarrow$
\\\hline

\textbf{NVS-WB}      & 24.78 &0.820 & 0.213 &  1.32\\

\textbf{GPNR}      & 24.11 & 0.793 & 0.255  & NA\\
\textbf{pixelNeRF}     & 20.43 &0.589 &0.550  & 5.30 \\
\textbf{Ours}      & \textbf{25.64}  &  \textbf{0.853}  & \textbf{0.148} & \textbf{0.11}\\

\hline

\end{tabular}

\caption{
 Application in generalizable gaussian splatting enabled by evolutive primitive organization: we show wide-baseline generalizable novel view synthesis from stereo images pairs on
the real-world RealEstate10k~\cite{re10k}  dataset. Our model outperform all baseline methods in terms PSNR, LPIPS, and SSIM, while requiring much less inference time.
}
\label{generalizable-gs}
\end{table}

\section{Limitations and Future Work}
 Although ERM has demonstrated broad applications ranging from enhancing performance of existing models to unlocking new possibilities for previously unattainable tasks, the current applications focus on the isolated utilization of individual evolutive elements among the three. Our current work reveals the substantial potential of evolutive rendering when applied to distinct components, however, we do not yet exploit its full potential. The integration of all of them remains a worthwhile avenue for future exploration and we will investigate this in the future.  Moreover, by enabling the differentiability of previous manually crafted components, additional parameters will be learned within the whole framework. Consequently, the incorporation of an evolutive element typically results in longer training time.

\section{Conclusion}

In this work, we introduce evolutive rendering models (ERMs) that replace the heuristic designs in rendering models with learnable components that are fully aligned with the final rendering objective.
In particular, we  introduce a comprehensive learning framework that underpins the evolution of three principal rendering elements, including
the gauge transformations, ray sampling mechanisms, and primitive organization. 
Our extensive experiments and thorough analysis show that the evolutive rendering models outperform their vanilla counterparts, hence demonstrating the large potential of evolutive rendering in computer graphics.

\bibliographystyle{acm}
\bibliography{reference}

\end{document}